\documentclass[11pt]{article}

\usepackage[final]{acl}

\usepackage{times}
\usepackage{latexsym}
\usepackage{amsmath}
\usepackage{amsfonts}
\usepackage{algorithm}
\usepackage{algpseudocode}
\usepackage[english]{babel}
\usepackage{inconsolata}
\usepackage{pgfplots}
\usepackage{pgfplotstable}
\usepackage{pifont} 
\usepackage{booktabs}
\usepackage{multirow}
\usepackage{adjustbox}
\usepackage{makecell}
\usepackage{float}
\usepackage{amssymb}
\usepackage{listings}
\usepackage{xcolor}
\usepackage{tcolorbox}
\usepackage[table]{xcolor}   
\usepackage{colortbl}
\tcbuselibrary{skins, breakable}
\usepackage{multicol} 
\usepackage{subcaption}
\usepackage{caption}
\pgfplotsset{compat=1.18}
\usepackage{fancyvrb}
\usepackage{enumitem}
\usepackage{hyperref}
\usepackage{array}
\usepgfplotslibrary{groupplots}

\usepackage{booktabs}

\usepackage{algorithm}
\usepackage{algpseudocode}
\usepackage{dsfont}
\usepackage{mathtools}
\usepackage{xspace}
\usepackage{tabularx}
\usepackage{arydshln} 

\usepackage[T1]{fontenc}

\usepackage[utf8]{inputenc}

\usepackage{microtype}

\usepackage{inconsolata}

\usepackage{graphicx}
\newcommand{\model}[1]{\texttt{#1}}
\title{Temp-R1: A Unified Autonomous Agent for Complex Temporal KGQA via Reverse Curriculum Reinforcement Learning}

\author{
  Zhaoyan Gong\textsuperscript{$\spadesuit$$\diamondsuit$},
  Zhiqiang Liu\textsuperscript{$\spadesuit$$\diamondsuit$},
  Songze Li\textsuperscript{$\spadesuit$$\diamondsuit$},
  Xiaoke Guo\textsuperscript{$\spadesuit$},
  Yuanxiang Liu\textsuperscript{$\spadesuit$},\\
  \textbf{Xinle Deng}\textsuperscript{$\spadesuit$},
  \textbf{Zhizhen Liu}\textsuperscript{$\clubsuit$},
  \textbf{Lei Liang}\textsuperscript{$\clubsuit$$\diamondsuit$},
  \textbf{Huajun Chen}\textsuperscript{$\spadesuit$$\diamondsuit$},
  \textbf{Wen Zhang}\textsuperscript{$\spadesuit$$\diamondsuit$\textdagger}
  \\
  \textsuperscript{$\spadesuit$} Zhejiang University
  \textsuperscript{$\clubsuit$} Ant Group \\
  \textsuperscript{$\diamondsuit$} ZJU-Ant Group Joint Lab of Knowledge Graph    \\
  \texttt{\{gongzhaoyan,zhang.wen\}@zju.edu.cn}
}

\begin{document}
\maketitle

\begingroup
  \renewcommand{\thefootnote}{\fnsymbol{footnote}}
  \setcounter{footnote}{0}
  \footnotetext{\textsuperscript{\textdagger} Corresponding author}
\endgroup

\begin{abstract}
Temporal Knowledge Graph Question Answering (TKGQA)  is inherently challenging, as it requires sophisticated reasoning over dynamic facts with multi-hop dependencies and complex temporal constraints. 
Existing methods rely on fixed workflows and expensive closed-source APIs, limiting flexibility and scalability. 
We propose \textbf{Temp-R1}, the first autonomous end-to-end agent for TKGQA trained through reinforcement learning. 
To address cognitive overload in single-action reasoning, we expand the action space with specialized internal actions alongside external action. To prevent shortcut learning on simple questions, we introduce reverse curriculum learning that trains on difficult questions first, forcing the development of sophisticated reasoning before transferring to easier cases. Our 8B-parameter Temp-R1 achieves state-of-the-art performance on \textsc{MultiTQ} and \textsc{TimelineKGQA}, improving 19.8\% over strong baselines on complex questions. Our work establishes a new paradigm for autonomous temporal reasoning agents. The code is available at \url{https://github.com/zjukg/Temp-R1}.
\end{abstract}

\definecolor{thinkblue}{RGB}{0, 0, 255}
\definecolor{searchcyan}{RGB}{0, 176, 240}
\definecolor{infobrown}{RGB}{192, 112, 0}
\definecolor{answerred}{RGB}{204, 0, 0}
\definecolor{planpurple}{RGB}{147, 39, 143}  
\definecolor{filterorange}{RGB}{247, 147, 30} 
\definecolor{rankgreen}{RGB}{0, 146, 69}     
\definecolor{bestcolor}{HTML}{E0F2F1}   
\definecolor{secondcolor}{HTML}{FFEBEE} 

\newcommand{\tplan}{{\color{planpurple}\texttt{<plan>}}\xspace}
\newcommand{\tthink}{{\color{thinkblue}\texttt{<think>}}\xspace}
\newcommand{\tsearch}{{\color{searchcyan}\texttt{<search>}}\xspace}
\newcommand{\tinfo}{{\color{infobrown}\texttt{<information>}}\xspace}
\newcommand{\tfilter}{{\color{filterorange}\texttt{<filter>}}\xspace}
\newcommand{\trank}{{\color{rankgreen}\texttt{<rank>}}\xspace}
\newcommand{\tanswer}{{\color{answerred}\texttt{<answer>}}\xspace}

\section{Introduction}

In the real world, knowledge is dynamic and constantly evolving. To prevent knowledge graphs~\cite{WOS:001356154700001,WOS:001356699700001} from storing outdated facts, Temporal Knowledge Graphs have emerged, consisting of quadruples in the form of \texttt{<subject, predicate, object, timestamp>}~\cite{jia2018tempquestions,saxena2021CronKGQA}.
 Consequently, Temporal Knowledge Graph Question Answering (TKGQA) requires reasoning over both entities and timestamps, making it significantly more challenging than conventional KGQA~\cite{saxena2020EmbedKGQA,enrich,skabench,layerlogitslogicempowering,liu2026cog}. 
Complex Temporal queries involve multi-hop reasoning, multiple constraints~\cite{su2024temporalsurvey}, 
and  multi-granular time~\cite{chen2023multiqa}, which typically require combining external knowledge retrieval with the temporal reasoning capabilities of large language models (LLMs)
~\cite{liu2025collaboration,zhang2025structured,guo2026}.

\begin{figure}[t] 
    \centering
    \includegraphics[width=\linewidth]{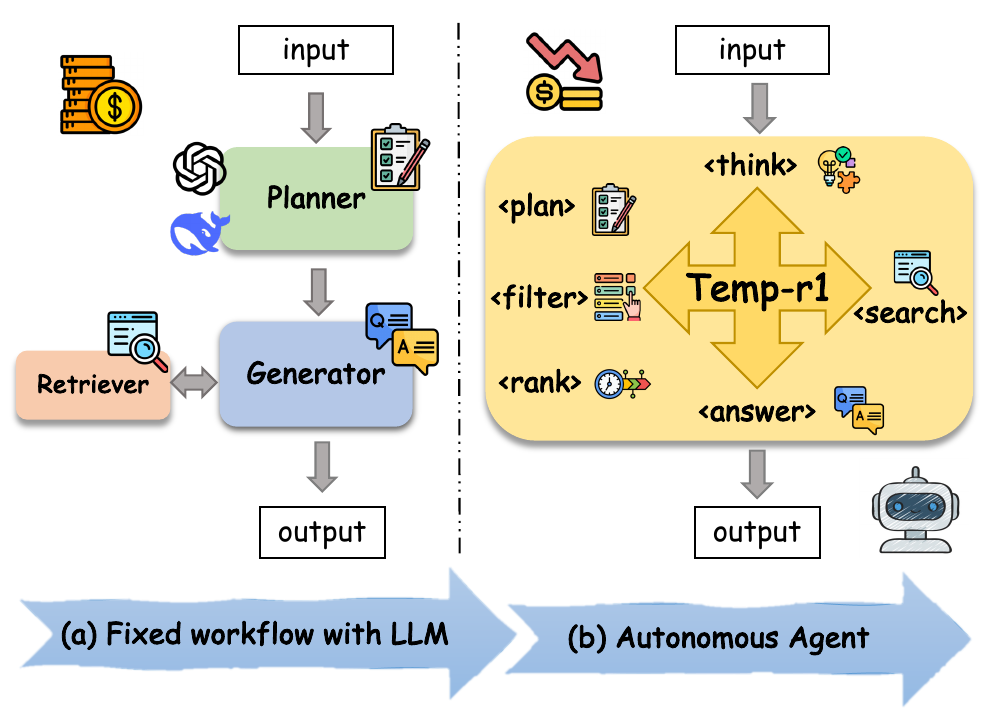} 
    \vspace{-1.2em}
    \caption{Paradigm shift in TKGQA systems. \textbf{Existing approaches} (a) suffer from fixed workflow and high API cost. \textbf{Temp-R1} (b) is an autonomous end-to-end agent trained for dynamic temporal reasoning.}
    \vspace{-1.2em}
    \label{fig:compare}
\end{figure}

Recent TKGQA methods have leveraged LLMs~\cite{chen2023ARI,gao2024GenTKGQA,jia2024faithful}. These approaches design fixed workflows comprising multiple carefully crafted prompts and components (decomposer, planner, generator), as illustrated in Figure~\ref{fig:compare}(a). While achieving promising results, they heavily rely on the capabilities of closed-source models, incurring high API costs, as shown in Table~\ref{tab:inference_cost} in Appendix~\ref{app:efficiency}. Moreover, such manually designed fixed workflows constrain the model's divergent thinking, leading to rigid solution strategies. The field urgently needs a flexible paradigm that can autonomously perform reasoning and invoke TKG retrieval tools, rather than adhering to predetermined workflows.

\begin{figure*}[t] 
    \centering
    \includegraphics[width=\linewidth]{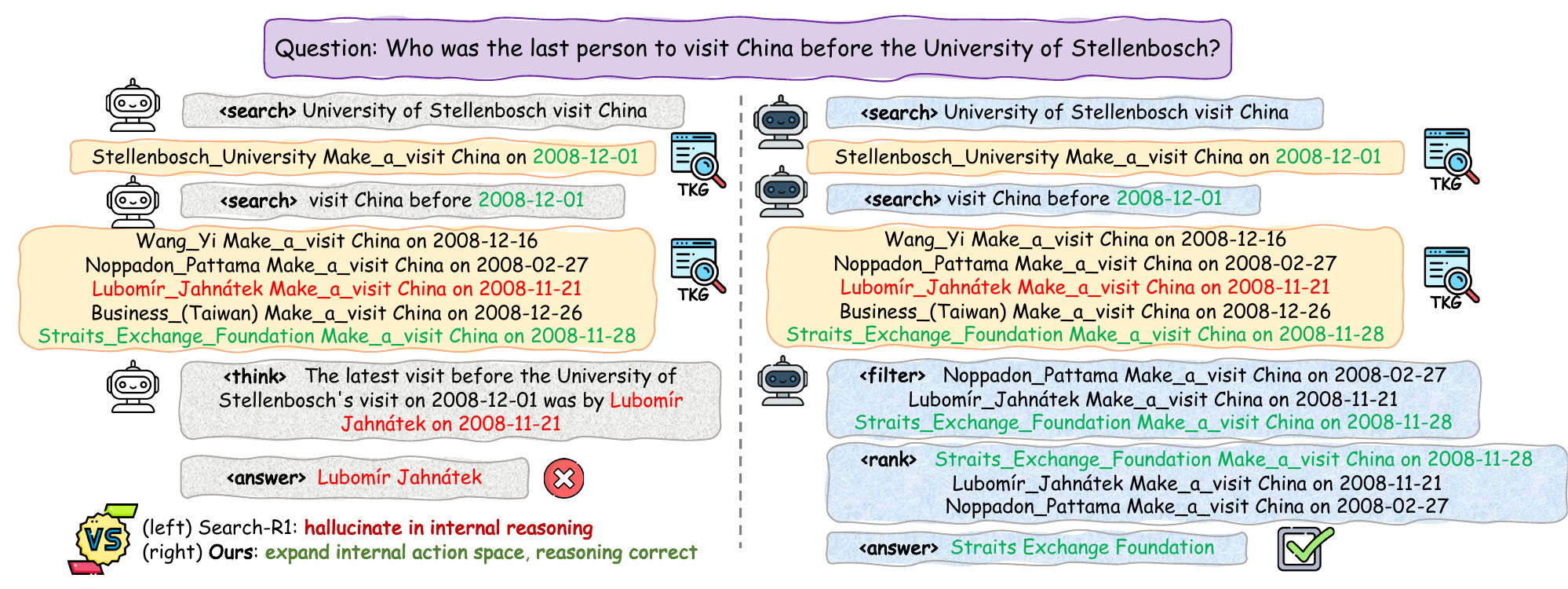} 
    \vspace{-1.5em}
    \caption{\textbf{Comparison of internal reasoning mechanisms: Temp-R1 vs. Search-R1.} By decoupling internal reasoning into explicit <filter> and <rank> actions (right), Temp-R1 maintains logical rigor and eliminates hallucinations in temporal sequencing. In contrast, the monolithic <think> block in Search-R1 (left) suffers from cognitive overload, leading to a failure in processing the retrieved temporal facts.}
    \vspace{-1.0em}
    \label{fig:actionspace}
\end{figure*}

The recent emergence of OpenAI o1~\cite{openai2024openaio1card} and DeepSeek-R1~\cite{guo2025deepseek-R1}, along with growing research on autonomous LLM agents~\cite{zheng2026predictexecutingmachinelearning,li2026whatsmissingscreentoactionuiintheloop}, offers new insights. Beyond their strong performance in mathematical reasoning and coding tasks, some methods combine reinforcement learning with search engine~\cite{jin2025search-R1,research}. 
However, applying these approaches to TKGQA exposes two critical challenges. 

\textbf{(1) Overloaded Internal Reasoning.}
Current search agents typically rely on a single internal reasoning tag, \tthink, which carries excessive cognitive load. For complex temporal questions, the agent must simultaneously handle autonomous search, strategic planning, semantic filtering, and chronological ranking.
Mixing these distinct cognitive demands within a single reasoning tag often leads to inadequate reinforcement learning and logical reasoning (as illustrated in Figure~\ref{fig:actionspace}).
\textbf{(2) The Shortcut Trap in Reinforcement Learning.}
Temporal QA datasets (e.g., \textsc{MultiTQ}) usually have uneven difficulty distributions. 
Applying trainset directly causes the agent to overfit to simpler questions and stop exploring harder reasoning paths once achieving high rewards on easy samples. Agents gradually develop path dependency, making it difficult to activate complex tool combinations.

To address these challenges, we propose \textbf{Temp-R1}, a flexible autonomous agent that self-explores diverse solution strategies. By training an 8B parameter model that outperforms closed-source models, we significantly reduce inference costs, as shown in Figure~\ref{fig:compare}(b).
To enhance the internal reasoning capabilities, we expand the action space. Temp-R1 not only retains the external \tsearch action but also decouples three internal reasoning actions from the \tthink token: \tplan enables initial problem analysis at rollout inception; \tfilter applies semantic relations and temporal constraints to information; and \trank orders filtered facts by timestamp. 
This compact yet expressive action design enables the agent to autonomously learn diverse reasoning trajectories through reinforcement learning~\cite{DBLP:journals/jair/KaelblingLM96}. 

To overcome shortcut behavior in reinforcement learning, we adopt a counter-intuitive approach: after a brief supervised fine-tuning (SFT) warm-up, we employ \textbf{reverse curriculum learning}, forcing the model to survive in high-difficulty environments first, compelling it to acquire sophisticated tool-chain logic before transferring to simpler environments, achieving a "dimensional reduction" effect that enhances generalization.

Experimental results demonstrate that Temp-R1 achieves superior overall performance across three datasets including \textsc{MultiTQ} and \textsc{TimelineKGQA}. Notably, on the challenging \texttt{Multiple} question type, our model achieves a significant \textbf{19.8\%} improvement over the strong baseline. Furthermore, our 8B parameter model establishes a new state-of-the-art overall performance, surpassing baselines built upon \model{GPT-4o}. Extensive ablation studies validate the effectiveness of reverse curriculum learning.
In summary, our contributions are as follows:

\begin{itemize}
    \item We propose a flexible autonomous agent for TKGQA, which decouples internal reasoning into an expanded action space, allowing for flexible and self-exploratory problem-solving.
    \item We introduce counter-intuitive reverse curriculum learning to mitigate the shortcut problem in reinforcement learning training.
    \item Extensive experiments show that our 8B model significantly outperforms powerful closed-source baselines, particularly on complex temporal reasoning tasks.
\end{itemize}

\section{Related Work}
\label{sec:related_work}

\subsection{Temporal Knowledge Graph QA}
Early approaches to TKGQA can be broadly categorized into three methods. 
Embedding-based methods encode questions and TKG quadruples into low-dimensional vectors, ranking answers based on vector semantic similarity~\cite{saxena2021CronKGQA,mavromatis2022tempoqr,chen2023multiqa,chen2022SubGTR}. 
Semantic parsing-based methods convert natural language questions into logical expressions to query TKGs, offering better interpretability~\cite{jia2018TEQUILA,neelam2021sygma,chen2024Prog-TQA,ding2022SF-TQA}. 
Additionally, several studies have incorporated Graph Neural Networks to capture structural dependencies~\cite{jia2021EXAQT,sharma2022TwiRGCN,liu2023LGQA,liu2023TMA}.

Recently, leveraging Large Language Models has become the mainstream paradigm. Some works focus on prompt engineering or fine-tuning: 
TimeR$^{4}$~\cite{qian2024timer4} explicitly reveals implicit time constraints through question rewriting. 
TempAgent~\cite{qianyihu-etal-2025-TempAgent} adapts the ReAct paradigms to the temporal domain, designing a toolkit with 10 specific tools. 
RTQA~\cite{gong-etal-2025-rtqa} employs a decomposition-based approach to solve sub-problems recursively from the bottom up. 
MemoTime~\cite{tan2025memotime} utilizes closed-source model APIs for reasoning and stores solution paths of similar questions as memory. 
PoK~\cite{qian2025poK} enhances TimeR$^{4}$ by generating plans for more comprehensive solution steps.
However, these methods follow fixed workflows, which limits flexibility and results in high API costs.

\subsection{Search Agent and Reinforcement Learning}
Reinforcement Learning has evolved from PPO~\cite{PPO} to more efficient direct optimization methods like DPO~\cite{DPO} and group-relative approaches like GRPO~\cite{GRPO}. Unlike SFT, which requires token-level supervision, RL-based models optimize policies using verifiable reward signals.

Notable models like OpenAI-o1~\cite{openai2024openaio1card} and DeepSeek-R1~\cite{guo2025deepseek-R1} have demonstrated exceptional performance in mathematical logic and coding ability~\cite{ARTIST,Zero-TIR}. In information retrieval, agents like Search-R1~\cite{jin2025search-R1,r1-searcher,research} apply these principles to autonomously search external knowledge.

\begin{figure*}[t] 
    \centering
    \includegraphics[width=\linewidth]{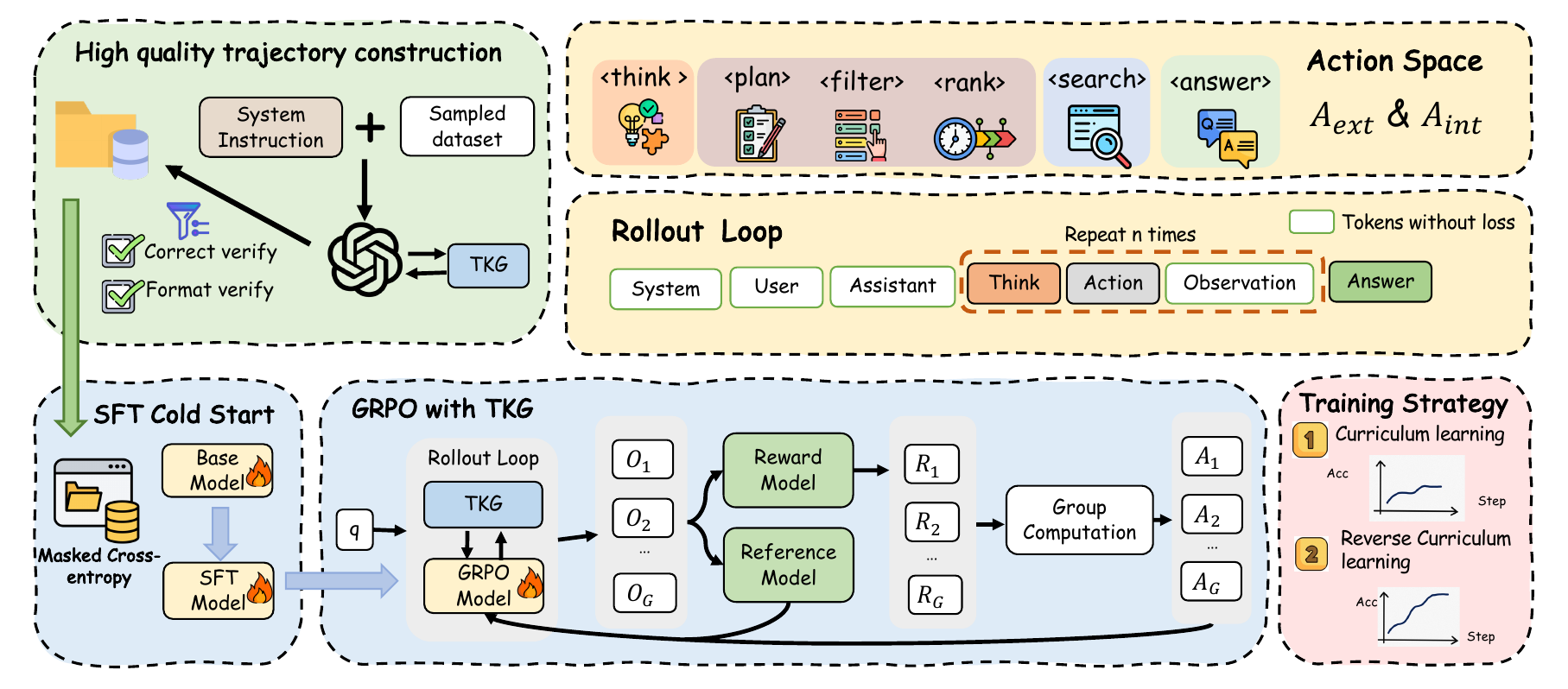}
    \vspace{-1.2em}
    \caption{\textbf{Overall architecture of Temp-R1.} The rollout loop features an expanded internal and external action space ($\mathcal{A}_{internal}$ and $\mathcal{A}_{external}$) designed for structured temporal reasoning. The model is trained via a two-stage process: supervised cold start from high-quality trajectories, followed by GRPO-based reinforcement learning supported by reverse curriculum learning strategies.}
    \vspace{-1.0em}
    \label{fig:training}
\end{figure*}

\section{Preliminary}
\label{sec:tkg_tkgqa}

\textbf{TKG.} A temporal knowledge graph \(\mathcal{G} = \{\mathcal{E}, \mathcal{P}, \mathcal{T}, \mathcal{F}\}\) is a directed graph where vertices are a set of entities \(\mathcal{E}\). The edges are a set of predicates \(\mathcal{P}\) with timestamps \(\mathcal{T}\). The quadruple set 
\(\mathcal{F} = \{(s, p, o, t) \mid \mathcal{E} \times \mathcal{P} \times \mathcal{E} \times \mathcal{T}\}\) 
represents the temporal facts, where \(s\) and \(o\) are subject and object, respectively, and \(p\) is the predicate between \(s\) and \(o\) at timestamp \(t\).

\paragraph{TKGQA.} TKGQA is a task to infer the correct answer to a natural language question \(q \in \mathcal{Q}\) based on relevant quadruples \(f = (s, p, o, t)\) in the temporal knowledge graph $\mathcal{G}$.
The answer can be an entity or a timestamp with varying granularities. 

\paragraph{Agent-based MDP Formulation.}
\label{sec:MDP}
Unlike conventional approaches, we formulate TKGQA as a Markov Decision Process (MDP) defined by the tuple $(\mathcal{S}, \mathcal{A}, \mathcal{P}, \mathcal{R})$. The state space $\mathcal{S}$ consists of states $s_t = (q, h_t)$ where $q$ is the original question and $h_t = [a_0, o_1, a_1, o_2, \ldots, a_{t-1}, o_t]$ represents the historical trajectory with actions $a_i$ and observations $o_i$. The action space $\mathcal{A}$ encompasses both internal reasoning actions and external tool invocations. The transition function $\mathcal{P}(s_{t+1} | s_t, a_t)$ operates as follows: when $a_t = \tsearch $, the next observation is $o_{t+1} = \text{Retriever}(a_t)$; when $a_t \in \{\tplan, \tfilter, \trank\}$, no external observation is generated ($o_{t+1} = \varnothing$); and when $a_t = \tanswer$, the episode terminates with $s_{t+1} = s_{\text{terminal}}$. The reward function $\mathcal{R}(s_t, a_t)$ will be elaborated in Section~\ref{sec:rl}. 

\section{Temp-R1}
\subsection{Overall Framework}
In this section, we present \textbf{Temp-R1}, a unified framework designed to internalize autonomous temporal reasoning abilities into compact language models. As illustrated in Figure~\ref{fig:training}, our methodology unfolds as a progressive learning pipeline, encompassing: (1) establishing a \textbf{rollout loop} with expanded action space; (2) performing a \textbf{supervised cold start} to teach basic formatting and action sequences; (3) conducting \textbf{Group Relative Policy Optimization} for discovering autonomous reasoning strategies; and (4) deploying a \textbf{reverse curriculum learning strategy} to prevent shortcut traps and path dependency.

\subsection{Rollout Loop of Temp-R1}
\label{sec:rollout}
Based on the MDP defined in Sec. \ref{sec:MDP}, the model explores the reasoning space through a structured rollout process. Unlike standard ReAct~\cite{YaoZYDSN023}, our rollout interleaves mandatory strategic planning with autonomous information distillation.
\paragraph{Action Space Expansion.}
Temp-R1 decomposes reasoning into explicit actions rather than relying on a single thinking token.
Internal actions include \tplan, \tfilter, and \trank for planning, filtering, and temporal organization, while the external action \tsearch retrieves knowledge.
The final action \tanswer terminates the rollout.
This decomposition exposes intermediate reasoning states and facilitates long-horizon policy optimization.


Formally, the action space is defined as:
\begin{equation}
    \mathcal{A} = \mathcal{A}_{\text{internal}} \cup \mathcal{A}_{\text{external}},
\end{equation} 
where internal actions are $\mathcal{A}_{\text{internal}} = \{\tplan, \tfilter, \trank\}$ and external actions are $\mathcal{A}_{\text{external}} = \{\tsearch\}$. The policy is modeled as
$
    a_t \sim \pi_\theta(a_t|s_t), 
$
where \(s_t\) denotes the textual context up to step \(t\) and \(a_t\) is one of the internal or external actions. The state is updated after each action as
$
s_{t+1} = (q, h_t \cup \{a_t, o_{t+1}\}), 
$
maintaining the complete interaction history for subsequent decision-making. 
The environment returns the corresponding observation such as retrieved information, which is then appended to the next input. 

\paragraph{Prompt Template.}
Table~\ref{tab:prompt_template1} shows the system prompt of Temp-R1, which specifies the action space, planning requirements, and output format. 
The process stops when \tanswer is produced or the maximum search turns $T_{\text{max}}$ are reached.

\subsection{Cold-Start: Learning Structured Format Alignment}
Although the rollout mechanism enables an expanded action space, a pretrained base model without task-specific alignment struggles to produce valid structured tags and coherent reasoning in the early stages of training.
To initialize a competent policy for reinforcement learning, we adopt a \textbf{Supervised Fine-Tuning (SFT)} phase as a cold start.

\paragraph{Data Construction.}
We construct a seed dataset $\mathcal{D}_{\text{sft}}$ consisting of $(q, \tau_{\text{gold}})$ pairs, where $\tau_{\text{gold}}$ contains mandatory planning and basic reasoning primitives leading to correct answers.
High quality trajectories are generated using \model{GPT-4o} and filtered to ensure structural correctness of reasoning tags and semantic correctness of the final answers.

\paragraph{Supervised Fine-Tuning.}
We train the model by minimizing the masked cross-entropy loss:
\begin{equation}
    \mathcal{L}_{\text{SFT}}(\theta) = - \frac{1}{T} \sum_{t=1}^{T} m_t \cdot \log \pi_\theta(x_t \mid x_{<t}),
\end{equation}
where $x$ denotes the token sequence and $m_t \in \{0,1\}$ is a selective loss mask.
Specifically, the loss is applied only to agent-generated tokens (e.g., \tplan, \tsearch, \tfilter, \tanswer), while system prompts, user inputs, and retrieved observations are masked out.
This selective token masking encourages the model to focus on structured reasoning and tool invocation, rather than memorizing retrieved content.

\subsection{RL: Acquiring Autonomous Reasoning and Strategy Exploration}
\label{sec:rl}
After SFT cold start, the model is further optimized with reinforcement learning to explore diverse reasoning strategies.

\paragraph{Group Relative Policy Optimization.}
Starting from the SFT-initialized policy $\pi_{\theta_{\mathrm{old}}}$, we adopt Group Relative Policy Optimization (GRPO) to refine decision making. 
The clipped surrogate objective with KL regularization is defined as:
\begin{equation}
\label{eq:grpo}
\begin{aligned}
    \mathcal{J}_{\mathrm{GRPO}}(\theta)
    = \frac{1}{G} \sum_{i=1}^{G}
      \Big[
      \min\big(
      \rho_i(\theta)\hat{A}_i,\,
      \mathrm{clip}(\rho_i(\theta),\\
      1-\epsilon,1+\epsilon)\hat{A}_i
      \big)
      - \beta\,\mathbb{D}_{\mathrm{KL}}(\pi_\theta\|\pi_{\mathrm{ref}})
      \Big],
\end{aligned}
\end{equation}
where $\rho_i(\theta)=\frac{\pi_\theta(\tau_i|q)}{\pi_{\theta_{\mathrm{old}}}(\tau_i|q)}$, $\epsilon$ is the clipping coefficient, and $\beta$ controls the KL penalty against the reference policy $\pi_{\mathrm{ref}}$.
For each query $q$, we sample $G$ trajectories $\{\tau_i\}_{i=1}^G$ from $\pi_\theta$, each receiving a terminal binary reward $r_i \in \{0,1\}$ based on answer correctness.
The group relative advantage is computed as
$
\hat{A}_i = \frac{r_i - \mathrm{mean}(\{r_k\})}{\mathrm{std}(\{r_k\}) + \eta}$,
where $\eta$ is a small constant for numerical stability.
This relative normalization encourages effective reasoning primitives and strategy exploration.

\begin{table}[t]
\centering
\small
\begin{tabular}{p{0.95\columnwidth}}
\toprule
\textbf{System Prompt} \\ \midrule
You are a question-answering assistant with a TKG.\\

\textbf{Start with planning}:

\tplan 

- Question type:\\
- Time constraints:\\
- Sub-questions:\\
- Answer format:\\
\texttt{{\color{planpurple}</plan>}}\\

\textbf{Then use tools as needed}: \\
-\tthink: Analytical reasoning. \texttt{{\color{thinkblue}</think>}}\\
-\tsearch\ Search the TKG. \texttt{{\color{searchcyan}</search>}}\\
-\tinfo\ Search results \texttt{{\color{infobrown}</information>}}. \\
-\tfilter\ Filter facts by temporal constraints. \texttt{{\color{filterorange}</filter>}}\\
-\trank\ Sort facts by date. \texttt{{\color{rankgreen}</rank>}}\\
-\tanswer\ Final answer \texttt{{\color{answerred}</answer>}} \\

\textbf{Important Constraints}: \\
1. Every response MUST start with \tplan. \\
2. Answer format must match the specification. \\
3. Only use facts from retrieved information. \\
4. End with \tanswer\ containing only the final answer. \\

Question: [Input Question] \\ \bottomrule
\end{tabular}
\caption{\textbf{System prompt} and action space template for Temp-R1. Different tags represent internal reasoning, external tools, and processing actions.}
\vspace{-1.0em}
\label{tab:prompt_template1}
\end{table}

\begin{table*}[t]
\centering

\small
\setlength{\tabcolsep}{8pt}
\begin{tabular}{l|cccc|cccc}
\toprule
\multirow{2}{*}{\textbf{Model}} &
\multicolumn{4}{c|}{\textbf{CronQuestion KG (In Domain)}} &
\multicolumn{4}{c}{\textbf{ICEWS Actor (Out of Domain)}} \\
\cmidrule(lr){2-5} \cmidrule(lr){6-9}
& \textbf{Overall} & \textbf{Simple} & \textbf{Medium} & \textbf{Complex}
& \textbf{Overall} & \textbf{Simple} & \textbf{Medium} & \textbf{Complex} \\
\midrule
RAG Baseline  & 0.235 & 0.704 & 0.092 & 0.009
              & 0.265 & 0.660 & 0.128 & 0.011 \\

LLaMA2-7B     & 0.169 & 0.049 & 0.143 & 0.282
              & 0.111 & 0.035 & 0.066 & 0.322 \\

GPT-4o        & 0.206 & 0.069 & 0.130 & 0.376
              & 0.113 & 0.051 & 0.035 & 0.353 \\

RTQA          & 0.298 & 0.608 & 0.218 & 0.135
              & - & - & - & - \\

PoK           & \cellcolor{secondcolor}\underline{0.651} & \cellcolor{secondcolor}\underline{0.737} & \cellcolor{bestcolor}\textbf{0.539} & \cellcolor{bestcolor}\textbf{0.683}
              & \cellcolor{secondcolor}\underline{0.602} & \cellcolor{secondcolor}\underline{0.744} & \cellcolor{bestcolor}\textbf{0.456} & \cellcolor{secondcolor}\underline{0.578} \\

\midrule
\textbf{Temp-R1}       & \cellcolor{bestcolor}\textbf{0.705} & \cellcolor{bestcolor}\textbf{0.960} & \cellcolor{secondcolor}\underline{0.486} & \cellcolor{secondcolor}\underline{0.672}
              & \cellcolor{bestcolor}\textbf{0.642} & \cellcolor{bestcolor}\textbf{0.866} & \cellcolor{secondcolor}\underline{0.388} & \cellcolor{bestcolor}\textbf{0.595} \\
\bottomrule
\vspace{-1.0em}
\end{tabular}
\caption{Performance comparison on \textsc{TimelineKGQA} across in-domain (CronQuestion KG) and out-of-domain (ICEWS Actor) scenarios. All results are evaluated using the \textbf{Hits@1} metric. The \colorbox{bestcolor}{\textbf{best}} and \colorbox{secondcolor}{\underline{second best}} scores under each metric are highlighted in colors, respectively. Baseline results are from~\cite{qian2025poK}.}
\label{tab:timelinekgqa_main}
\end{table*}

\paragraph{Reward Function.}
Temp-R1 is trained with a rule-based terminal reward that directly evaluates final answer correctness.
The reward is defined as $R = 1$ if $a_{\mathrm{pred}} = a_{\mathrm{gold}}$, and $R = 0$ otherwise, where $a_{\mathrm{pred}}$ denotes the extracted final answer and $a_{\mathrm{gold}}$ is the ground truth.

\begin{table}[t]

\centering
\small
\setlength{\tabcolsep}{3.5pt}
\begin{adjustbox}{max width=\textwidth}

\begin{tabular}{lccccc}
\toprule
\multirow{2}{*}{\textbf{Model}} &
\multirow{2}{*}{\textbf{Overall}}  &
\multicolumn{2}{c}{\textbf{Question Type}} &
\multicolumn{2}{c}{\textbf{Answer Type}} \\
\cmidrule(lr){3-4} \cmidrule(lr){5-6}
& & \textbf{multiple} & \textbf{single} & \textbf{entity} & \textbf{time} \\
\midrule

\multicolumn{6}{c}{\textbf{TKG Embedding-based method}} \\
\midrule
EmbedKGQA & 0.206 & 0.134 & 0.235 & 0.290 & 0.001 \\
CronKGQA & 0.279 & 0.134 & 0.337 & 0.328 & 0.156 \\
MultiQA  & 0.293 & 0.159 & 0.347 & 0.349 & 0.157 \\
\midrule

\multicolumn{6}{c}{\textbf{Prompt-based LLM}} \\
\midrule
ARI & 0.380 & 0.210 & 0.680 & 0.394 & 0.344 \\
TempAgent & 0.702 & 0.316 & 0.857 & 0.624 & 0.870 \\
MemoTime & 0.730 & \cellcolor{secondcolor}\underline{0.459} & 0.829 & 0.677 & 0.846 \\
RTQA & 0.765 & 0.424 & 0.902 & 0.692 & 0.942 \\
\midrule

\multicolumn{6}{c}{\textbf{FineTune-based LLM}} \\
\midrule
Search-R1 & 0.352 & 0.094 & 0.474 & 0.230 & 0.705 \\
TimeR$^{4}$ & 0.728 & 0.335 & 0.887 & 0.639 & 0.945 \\
PoK & \cellcolor{secondcolor}\underline{0.779} & 0.409 & \cellcolor{bestcolor}\textbf{0.929} & \cellcolor{secondcolor}\underline{0.696} & \cellcolor{secondcolor}\underline{0.962} \\
\midrule
\textbf{Temp-R1} &
\cellcolor{bestcolor}\textbf{0.780} & 
\cellcolor{bestcolor}\textbf{0.550} &
\cellcolor{secondcolor}\underline{0.888} &
\cellcolor{bestcolor}\textbf{0.714} &
\cellcolor{bestcolor}\textbf{0.969} \\
\bottomrule
\end{tabular}
\end{adjustbox}
\caption{Performance comparison on the \textsc{MultiTQ} test set. Evaluation results are reported using the \textbf{Hits@1} metric across various question and answer categories. The \colorbox{bestcolor}{\textbf{best}} and \colorbox{secondcolor}{\underline{second best}} scores under each metric are highlighted in colors, respectively.}
\label{tab:multitq_main}
\vspace{-0.5em}
\end{table}

\subsection{Reverse Curriculum Learning Strategy}
Traditional curriculum learning progresses from easy to hard ($\mathcal{D}_{\text{easy}} \rightarrow \mathcal{D}_{\text{medium}} \rightarrow \mathcal{D}_{\text{hard}}$), but this suffers from a shortcut trap in TKGQA tasks where models learn minimal patterns like $\tsearch \rightarrow \tanswer$ that suffice for simple questions but fail to activate complex tool combinations needed for difficult temporal reasoning.
Figure~\ref{fig:noRCL} empirically demonstrates that, without Reverse Curriculum Learning, training is dominated by path dependency and converges prematurely.

To overcome this, Temp-R1 employs a reverse curriculum strategy. We prioritize complex multi-hop queries $\mathcal{D}_\text{multi}$ to force the model to master sophisticated temporal reasoning and tool combinations first. Simpler queries $\mathcal{D}_\text{single}$ are introduced only after a warm-up threshold $T_0$:
\begin{equation}
\mathcal{D}_t=
\begin{cases}
\mathcal{D}_\text{multi}, & t\le T_0,\\
\mathcal{D}_\text{multi}\cup \mathcal{D}_\text{single}, & t>T_0.
\end{cases}
\end{equation}
This approach ensures that high-level reasoning capabilities are established early, which then generalize robustly to simpler tasks.
\begin{table}[t]

\centering
\small
\setlength{\tabcolsep}{3pt}
\begin{tabular}{lccccc}
\toprule
\multirow{2}{*}{\textbf{Model}} 
& \multirow{2}{*}{\textbf{Overall}} 
& \multicolumn{2}{c}{\textbf{Question Type}} 
& \multicolumn{2}{c}{\textbf{Answer Type}} \\
\cmidrule{3-6}
& & \textbf{multiple} & \textbf{single} & \textbf{entity} & \textbf{time} \\
\midrule
\textbf{Temp-R1} 
& \textbf{0.780} & \textbf{0.550} & \textbf{0.888} & \textbf{0.714} & \textbf{0.969} \\
w/o $\mathcal{A}_{int}$ 
& 0.620 & 0.388 & 0.729 & 0.563 & 0.783 \\
w/o Reverse CL
& 0.556 & 0.143 & 0.750 & 0.447 & 0.868 \\
w/o SFT
& 0.582 & 0.325 & 0.703 & 0.536 & 0.713 \\
\bottomrule
\end{tabular}
\caption{\textbf{Ablation study} on \textsc{MultiTQ} using Hits@1. We evaluate the contribution of each key component by removing the internal actions ($\mathcal{A}_{int}$), the Reverse Curriculum Learning strategy, and the SFT cold-start.}
\label{tab:ablation}
\end{table}

\section{Experiments}
\subsection{Experimental Setups}
\paragraph{Datasets.}
We train Temp-R1 on two datasets: \textsc{MultiTQ} and \textsc{TimelineKGQA-Cron}. 
We evaluate Temp-R1 on three benchmarks: \textsc{MultiTQ}, \textsc{TimelineKGQA-cron} and \textsc{TimelineKGQA-icews-actor}. The \textsc{TimelineKGQA-ICEWS-Actor} dataset serves as an out-of-domain benchmark to evaluate the generalization ability of Temp-R1.
Detailed statistics and category distributions are provided in Appendix~\ref{app:dataset}.

\paragraph{Baselines.}
We compare Temp-R1 against three types of baselines on \textsc{MultiTQ}: (1) \textbf{TKG Embedding-based methods}, including EmbedKGQA~\cite{saxena2020EmbedKGQA}, CronKGQA~\cite{saxena2021CronKGQA}, and MultiQA~\cite{chen2023multiqa}; (2) \textbf{LLM with prompt engineering}, including ARI~\cite{chen2023ARI}, TempAgent~\cite{qianyihu-etal-2025-TempAgent}, MemoTime~\cite{tan2025memotime}, and RTQA~\cite{gong-etal-2025-rtqa}; (3) \textbf{LLM with Fine-Tuning}, including Search-R1~\cite{jin2025search-R1}, TimeR$^{4}$~\cite{qian2024timer4}, and PoK~\cite{qian2025poK}.
For \textsc{TimelineKGQA-cron} and \textsc{TimelineKGQA-icews-actor}, due to its complexity, existing embedding-based models are not directly applicable. We adopt five strong baselines. More baseline details are provided in Appendix~\ref{app:baseline}.

\begin{figure}[t] 
    \centering
    \includegraphics[width=\linewidth]{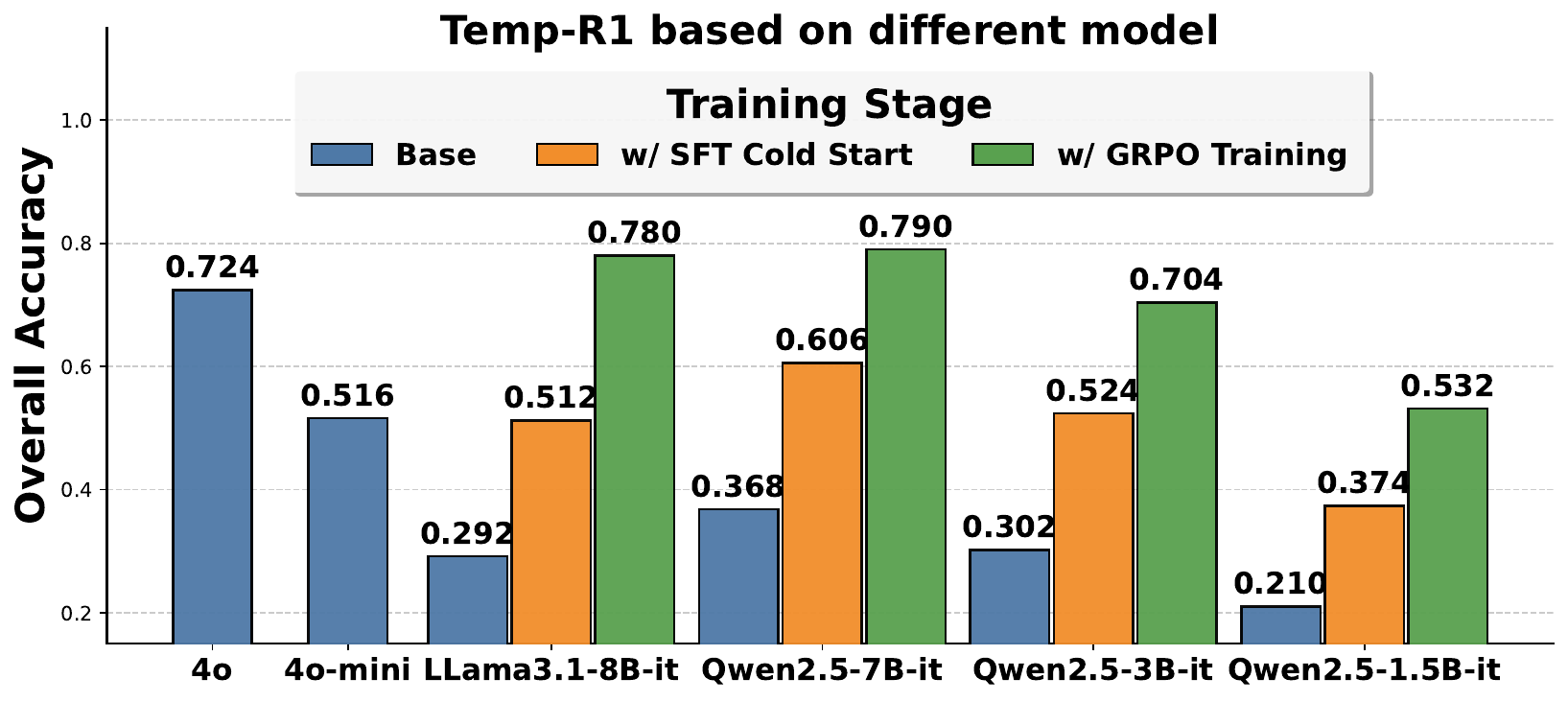} 
    \vspace{-1.0em}
    \caption{Performance comparison of Temp-R1 across different \textbf{backbone models} (\model{GPT-4o}, \model{GPT-4o-mini}, \model{LLaMA3.1-8B-it}, \model{Qwen2.5} series) and \textbf{training stages} (base model, w/SFT cold start, w/GRPO training).}
    \vspace{-0.5em}
    \label{fig:different_backbone-all}
\end{figure}

\paragraph{Implementation Details.}

We fine-tune the \model{Llama3.1-8B-Instruct} model, employing the same E5 retriever followed by~\cite{jin2025search-R1}. 
The SFT dataset consists of approximately 1,000 high-quality trajectories. The GRPO stage uses unlabeled QA pairs. Since the \textsc{MultiTQ} trainset is considerably large, we utilize only 9\% of it for RL training. Detailed settings for both SFT and GRPO training config can be found in the Appendix~\ref{app:details}.

\begin{figure}[t] 
    \centering
    \includegraphics[width=\linewidth]{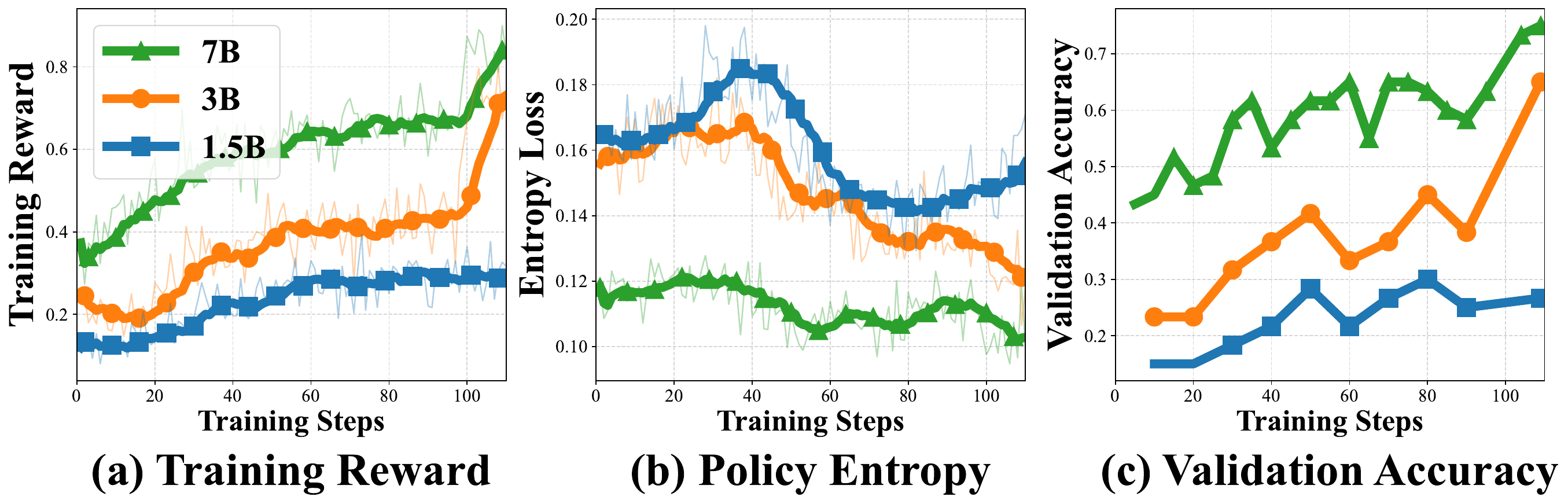} 
    \vspace{-1.5em}
    \caption{Training dynamics across different Qwen2.5 \textbf{backbone sizes} (7B, 3B, 1.5B). (a) Training Reward, (b) Policy Entropy, and (c) Val Accuracy}
    \label{fig:qwen2.5 seris}
\end{figure}

\begin{figure}[t] 
    \centering
    \includegraphics[width=\linewidth]{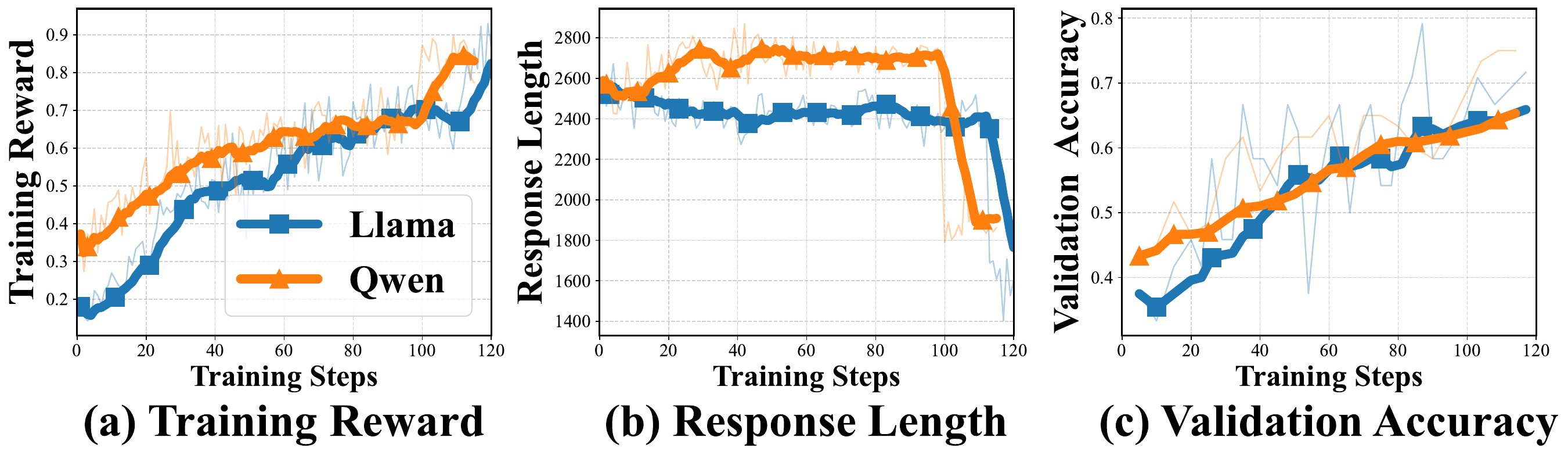} 
    \vspace{-1.5em}
    \caption{Training dynamics between LLaMA and Qwen \textbf{Architecture}. (a) Training Reward, (b) Response Length, and (c) Val Accuracy.}
    \vspace{-1.0em}
    \label{fig:qwen vs llama}
\end{figure}

\begin{figure}[t] 
    \centering
    \includegraphics[width=\linewidth]{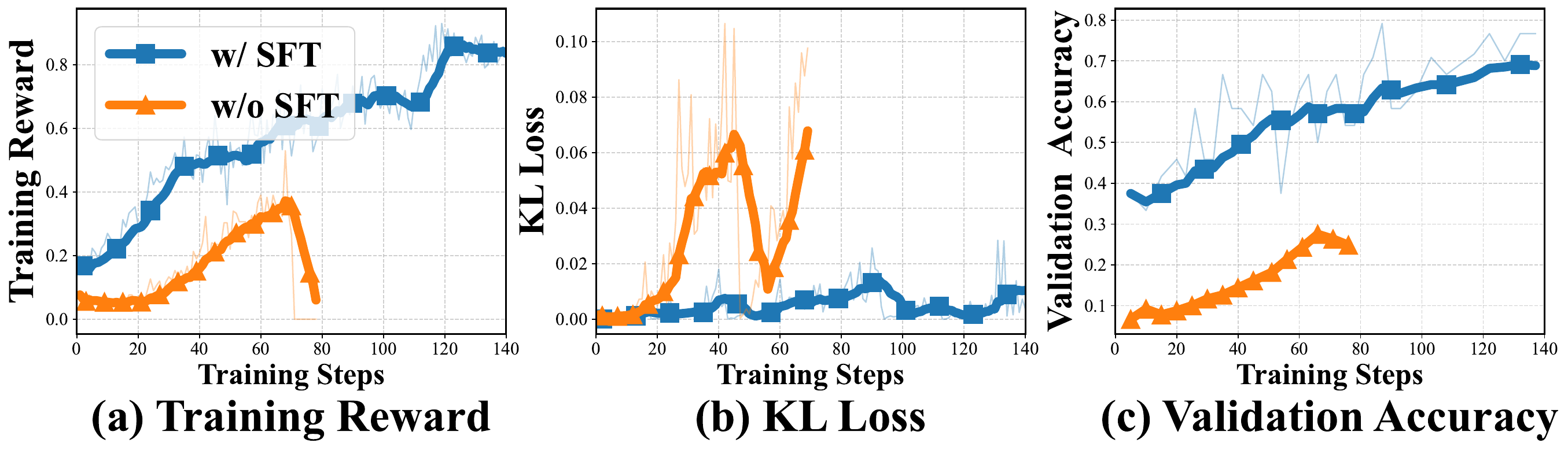} 
    \vspace{-1.5em}
    \caption{Training dynamics with/without \textbf{SFT Cold Start}. (a) Training Reward, (b) KL Loss, and (c) Val Accuracy. The blue line shows better stability and higher performance compared to the orange line.}
    \label{fig:noSFT}
\end{figure}

\begin{figure}[t] 
    \centering
    \includegraphics[width=\linewidth]{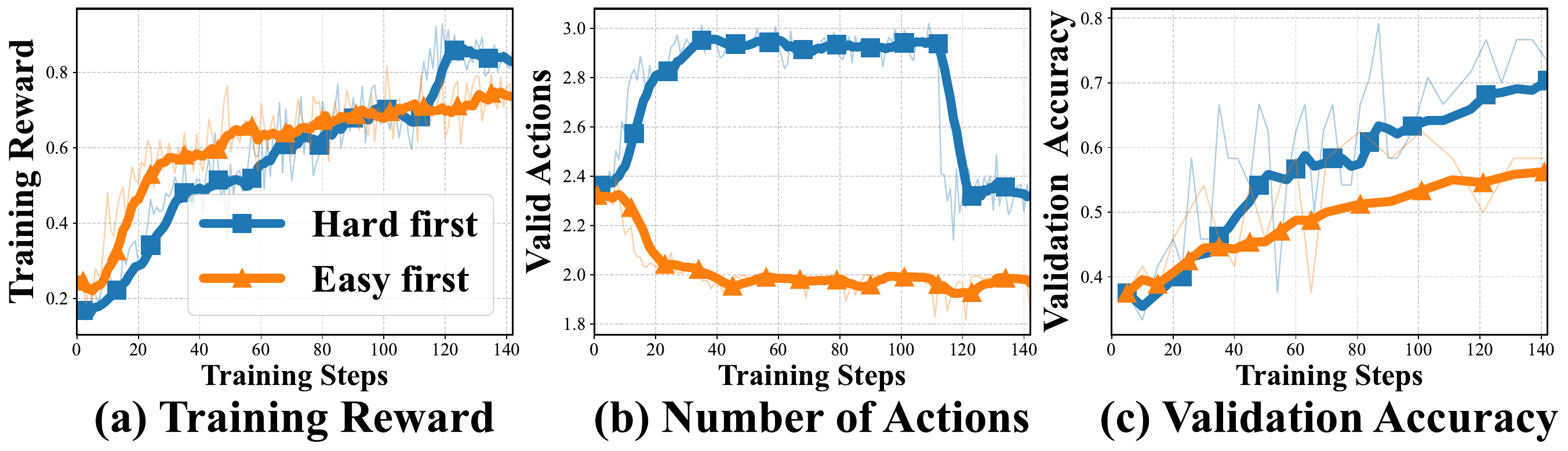} 
    \vspace{-1.5em}
    \caption{Training dynamics with/without \textbf{Reverse Curriculum Learning}. (a) Training Reward, (b) Number of Actions, (c) Val Accuracy. Without Reverse Curriculum Learning, all metrics show path dependency.}
    \vspace{-1.0em}
    \label{fig:noRCL}
\end{figure}

\subsection{Main Results}
\label{subsec:main_results}

The main experimental results on \textsc{MultiTQ} and \textsc{TimelineKGQA} are summarized in Table~\ref{tab:multitq_main} and Table~\ref{tab:timelinekgqa_main}, respectively. Our proposed Temp-R1 outperforms all baseline methods across most metrics.

\paragraph{Overall Superiority and Model Efficiency.}
As shown in Table~\ref{tab:multitq_main}, Temp-R1 achieves the state-of-the-art performance with an overall score of 0.780 on \textsc{MultiTQ}. It is worth noting that while baselines rely on powerful closed-source LLMs (e.g., \model{GPT-4o-mini}, or \model{DeepSeek-V3}), our Temp-R1, based on an 8B open-source backbone, achieves superior results. This demonstrates that a smaller model can surpass the performance of much larger, proprietary models that use fixed workflows.

\paragraph{Handling Complex Temporal Reasoning.} 
A key strength of Temp-R1 is its ability to handle sophisticated temporal questions involving multi-hop reasoning and multiple constraints. \noindent \textbf{(1) Multiple Constraints:} In the \texttt{multiple} category of \textsc{MultiTQ} (Table~\ref{tab:multitq_main}), Temp-R1 achieves a score of 0.550, outperforming the state-of-the-art model by 9.1\%. This improvement underscores its superior capacity for resolving intricate constraints that often challenge existing models. \noindent \textbf{(2) Multi-granular Precision:} The model demonstrates robust accuracy across diverse answer types, specifically reaching 0.969 on \texttt{time} category queries. This reflects high precision in pinpointing exact dates and intervals across various temporal granularities.

\paragraph{Generalization to Out-of-Domain Scenarios.}
Table~\ref{tab:timelinekgqa_main} highlights the robustness of Temp-R1 when facing domain shifts. On the \textsc{Timeline-ICEWS-Actor} dataset, which serves as an Out-of-Domain testbed, Temp-R1 maintains its lead with an overall score of 0.642, surpassing \textit{PoK} (0.602). While frozen models like \model{GPT-4o} struggle significantly in this specialized temporal domain (only 0.113), Temp-R1 exhibits remarkable stability.

\begin{figure*}[t] 
    \centering
    \includegraphics[width=\textwidth]{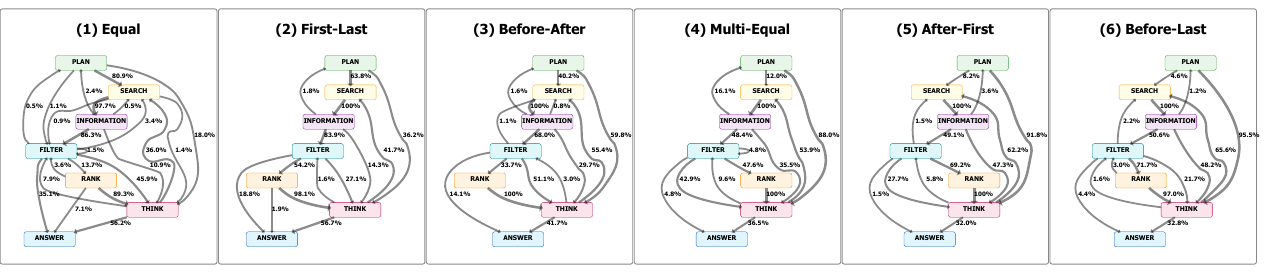} 
    \vspace{-1.0em}
    \caption{\textbf{Autonomous and flexible reasoning trajectories}: state transition diagrams across six question types.}
    \label{fig:6 type tag}
\end{figure*}

\subsection{Ablation Study}

We evaluate the contribution of each component in \textbf{Temp-R1} via ablation experiments on \textsc{MultiTQ}, with results summarized in Table~\ref{tab:ablation}.

\paragraph{Effect of Internal Reasoning Actions.} 
Removing $\mathcal{A}_{int}$ causes overall accuracy to drop from 0.780 to 0.620, with \texttt{Multiple} performance declining from 0.550 to 0.388. When only external actions are available, the model must perform all temporal reasoning implicitly in the \tthink tag, which increases cognitive load and often leads to reasoning errors or even hallucinated conclusions.

\paragraph{Significance of SFT Cold Start.} 
Removing SFT initialization reduces overall performance to 0.582, with time-type accuracy dropping from 0.969 to 0.713. 
Without the supervised fine‑tuning initialization, the model starts reinforcement learning directly on challenging temporal tasks and fails to learn consistent reasoning formats. 
SFT provides essential prior knowledge of temporal patterns and formatting, serving as a bridge between static KG understanding and dynamic policy training.

\paragraph{Role of Reverse Curriculum Learning.}
Removing reverse curriculum learning reduces the overall score to 0.556.
More critically, performance on hard queries drops sharply from 0.550 to 0.143, indicating that the agent overfits to simple questions and fails to solve complex reasoning tasks.

\subsection{Impact of Backbone Model Selection}
\paragraph{Model Scale: 7B, 3B vs. 1.5B.}

Figure~\ref{fig:qwen2.5 seris} confirms that performance scales with parameter size. While all scales show steady improvement during training, the 7B model achieves peak accuracy of 0.790 versus 0.532 for the 1.5B variant. These results show that model capacity is critical for complex reasoning, while validating that our training strategy remains effective across different scales.

\paragraph{Architecture Comparison: Llama vs. Qwen.}
Figure~\ref{fig:qwen vs llama} demonstrates the broad adaptability of Temp-R1 across Llama and Qwen architectures. Both backbones exhibit stable training and comparable performance (0.790 for Qwen vs. 0.780 for Llama), with Qwen showing slightly longer response trajectories (Figure~\ref{fig:qwen vs llama}b). This confirms that Temp-R1 effectively enhances reasoning capabilities independent of underlying architecture.

\begin{figure}[t] 
    \centering
    \includegraphics[width=\linewidth]{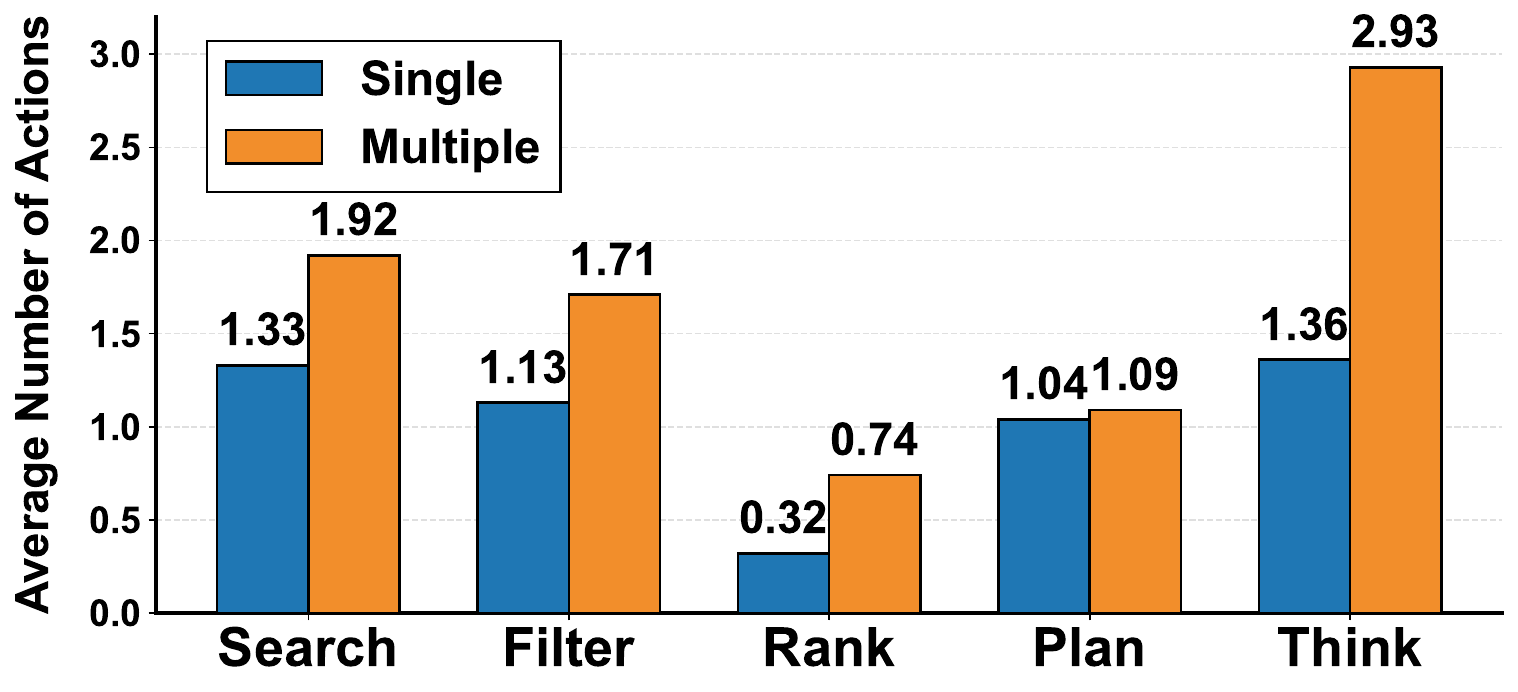} 
    \caption{Comparison of the \textbf{average number of actions} for \texttt{Single} vs. \texttt{Multiple} tasks across internal action, external action and thinking.}
    \vspace{-0.5em}
    \label{fig:action number}
\end{figure}

\subsection{Analysis of Training Strategy Choices}

\paragraph{The Role of Cold Start: w/ SFT vs. w/o SFT.}
Figure~\ref{fig:noSFT} highlights SFT as a prerequisite for RL stability. Without SFT, the model suffers from early training collapse and uncontrollable KL loss spikes, leading to poor validation accuracy. SFT provides a high-quality initial policy, ensuring stable optimization and preventing the model from deviating into incoherent reasoning.

\paragraph{Effectiveness of GRPO Optimization: w/ GRPO vs. w/o GRPO.}

Figure~\ref{fig:different_backbone-all} confirms the effectiveness of GRPO. Across all backbones, the "w/ GRPO Training" stage yields substantial accuracy gains over "Base" and "w/ SFT Cold Start" stages. This demonstrates that RL refinement is crucial for internalizing complex constraints and optimizing reasoning trajectories beyond imitation learning.

\paragraph{Curriculum Learning: Easy First vs. Hard First.}

Figure~\ref{fig:noRCL} highlights the advantages of reverse curriculum learning (\textit{Hard First}). While the \textit{Easy First} approach offers rapid initial reward growth, it often plateaus by relying on simplistic shortcuts. Conversely, the \textit{Hard First} strategy promotes extensive exploration and higher action complexity (Figure~\ref{fig:noRCL}b), avoiding local optima and achieving superior validation accuracy and generalization.

\subsection{Reasoning Trajectory Analysis}

\paragraph{Dynamics of Action State Transitions.}

As shown in Figure~\ref{fig:6 type tag}, Temp-R1 operates as a flexible agent rather than a rigid workflow. Transition probabilities vary notably across problem types, indicating adaptive trajectory planning. The model dynamically selects optimal action sequences according to query demands. Detailed case analysis is provided in Appendix~\ref{app:case study} and Table~\ref{tab:case_study_before/after}--\ref{tab:case_study_after-first}.

\paragraph{Reasoning Complexity: Number of Actions.}
Figure~\ref{fig:action number} shows that the reasoning trajectory length scales with task difficulty. Compared to simple queries, complex problems trigger more internal actions, where \tthink increase from 1.36 to 2.93 and \tsearch calls rise from 1.33 to 1.92. This adaptive behavior shows that the agent can autonomously devote more reasoning steps and tool calls when solving complex problems. Efficiency analysis is provided in Appendix~\ref{app:efficiency} and Table~\ref{tab:training_cost}--~\ref{tab:inference_cost}.

\section{Conclusion}
In this paper, we introduced \textbf{Temp-R1}, a novel autonomous agent framework that transforms TKGQA from a fixed-workflow paradigm into a flexible, self-exploratory reasoning process. Through expanded action space design and reverse curriculum learning, our 8B model achieves state-of-the-art performance, substantially outperforming \model{GPT-4o}-based methods. This work establish a new paradigm for building cost-effective, flexible autonomous reasoning agents.

\section*{Limitations}
Due to computational constraints, our experiments are limited to models with parameter scales up to 8B. We did not conduct fine‑tuning or reinforcement learning training on larger backbone models (such as 14B or beyond), which might further enhance temporal reasoning capability. 
Therefore, the scalability of Temp‑R1 to higher‑capacity models remains to be empirically verified. 

Furthermore, while we innovatively integrated reverse curriculum learning into the TKGQA task and observed better performance gains than curriculum learning, these results do not necessarily imply that it is a universal solution for all reasoning tasks or data distributions. Our findings primarily demonstrate its efficacy within the TKGQA task, and its generalizability to broader, non-temporal domains requires further investigation.

\section*{Ethical considerations}
In this paper, we investigate temporal knowledge graph question answering (TKGQA), focusing on complex reasoning over structured temporal data. Our method is developed and evaluated using publicly available and widely used datasets, including \textsc{MultiTQ} and \textsc{TimelineKGQA}. These datasets are constructed from open sources and do not contain any sensitive or personally identifiable information. Therefore, we believe that our work does not pose any ethical concerns.

\section*{Acknowledgements}
This work is founded by National Natural Science Foundation of China (NSFC62306276/
NSFCU23B2055), New Generation Artificial Intelligence-National Science and Technology Major Project 2030 (2025ZD0122800), Yongjiang Talent Introduction Programme (2022A-238-G), and Fundamental Research Funds for the Central Universities (226-2023-00138). This work was supported by Ant Group. 

\bibliography{custom}

\clearpage
\appendix
\label{sec:appendix}
\section{Datasets \& Corpora}
\label{app:dataset}

We evaluate on three TKGQA benchmarks with complementary characteristics. \textsc{MultiTQ} offers large-scale QA pairs with diverse temporal granularities, while \textsc{TimelineKGQA} covers questions with varying complexity and time formats. Table~\ref{tab:data_splits} shows the detailed statistics.

\paragraph{MULTITQ.}
Built on ICEWS-Event with 461K temporal facts, \textsc{MultiTQ} contains 386K/58K/55K train/val/test questions. Questions are categorized as \texttt{Single} (73\%, simple factual lookup or single-hop reasoning) or \texttt{Multiple} (27\%, complex multi-hop reasoning with temporal ordering and constraints). The knowledge graph uses quadruples (subject, predicate, object, timestamp). Questions cover Before/After temporal sequencing, First/Last comparative reasoning, and both explicit/implicit temporal constraints across multiple granularities (day/month/year/entity). The imbalanced complexity distribution makes it ideal for studying curriculum learning and reward hacking in RL settings.
\begin{table*}[htbp]
    \centering
    
    \small
    \setlength{\tabcolsep}{6pt}
    \begin{tabular}{lllrrr}
        \toprule
        \multirow{2}{*}{\textbf{Dataset}} & \multirow{2}{*}{\textbf{TKG}} & \multicolumn{4}{c}{\textbf{Question}}\\
        \cmidrule(lr){3-6}
        & & \textbf{Type} & \textbf{Train} & \textbf{Val} & \textbf{Test} \\
        \midrule

        \multirow{3}{*}{\textbf{MultiTQ}} & \multirow{3}{*}{461,329}  
            & Single   & 283,482 & 41,735 & 38,864 \\
            &  & Multiple & 103,305 & 16,244 & 15,720 \\ 
        \cmidrule(lr){3-6}
            &  & \textbf{Total}    & 386,787 & 57,979 & 54,584 \\
        \midrule
        
        \multirow{4}{*}{\textbf{Timeline-ICEWS-Actor}} & \multirow{4}{*}{108,005}  
            & Simple  & 17,982 & 5,994 & 5,994 \\
            & & Medium  & 15,990 & 5,330 & 5,330 \\
            & & Complex & 19,652 & 6,550 & 6,550 \\ 
        \cmidrule(lr){3-6}
            & & \textbf{Total}   & 53,624 & 17,874 & 17,874 \\
        \midrule
        
        \multirow{4}{*}{\textbf{Timeline-CronQuestion}} & \multirow{4}{*}{328,635} 
            & Simple  & 7,200 & 2,400 & 2,400 \\
            & & Medium  & 8,252 & 2,751 & 2,751 \\
            & & Complex & 9,580 & 3,193 & 3,193 \\ 
        \cmidrule(lr){3-6}
            & & \textbf{Total}   & 25,032 & 8,344 & 8,344 \\
            
        \bottomrule
    \end{tabular}
    \caption{Detailed statistics of the TKGQA datasets. Questions are categorized by difficulty or type, and the background TKG size is provided for each.}
    \label{tab:data_splits}
\end{table*}
\paragraph{Timeline-ICEWS-Actor.}
Constructed from ICEWS political events with 108K facts, this dataset provides 54K/18K/18K train/val/test questions stratified into three balanced difficulty levels: \texttt{Simple} (33\%, direct retrieval), \texttt{Medium} (30\%, single-hop reasoning), and \texttt{Complex} (37\%, multi-hop reasoning). The domain-specific focus on international relations and balanced difficulty distribution enable controlled evaluation of progressive learning strategies.

\paragraph{Timeline-CronQuestion.}
Derived from CronQuestions TKG with 329K facts and 25K/8K/8K train/val/test splits, this dataset emphasizes temporal intervals and duration reasoning. Questions span three complexity levels: \texttt{Simple} (29\%), \texttt{Medium} (33\%), and \texttt{Complex} (38\%). Unlike the other datasets, it requires temporal arithmetic, interval calculations, and overlapping period reasoning, with answers including durations like ``3 years''.

These datasets complement each other in scale, complexity distribution, and reasoning requirements, enabling comprehensive evaluation from simple retrieval to complex temporal inference.

\section{Baselines}
\label{app:baseline}

We compare Temp-R1 against a comprehensive set of baselines covering different paradigms in temporal knowledge graph question answering (TKGQA). Specifically, we evaluate three categories of methods on \textsc{MultiTQ}, and five state-of-the-art baselines on \textsc{Timeline-ICEWS-Actor} and \textsc{Timeline-CronQuestion}. Our baseline selection represents the most recent and competitive approaches in the field, presenting significant challenges for evaluation.

\noindent\textbf{(1) TKG Embedding-based Methods.}
These approaches learn low-dimensional temporal embeddings for entities and relations in TKGs, performing reasoning by computing similarity scores between question representations and candidate facts. 
Representative models include EmbedKGQA~\cite{saxena2020EmbedKGQA}, which extends static KG embedding techniques to question answering; 
CronKGQA~\cite{saxena2021CronKGQA}, which integrates temporal intervals into entity and relation embeddings; and 
MultiQA~\cite{chen2023multiqa}, which aggregates multi-granular temporal information for reasoning.

\noindent\textbf{(2) LLM with Prompt Engineering.}
This category employs large language models as zero-shot or few-shot reasoners guided by carefully designed prompts. These methods represent particularly \textit{strong baselines} as they leverage powerful closed-source model APIs, demonstrating capabilities fundamentally different from our approach.
Representative examples include:
ARI~\cite{chen2023ARI} augments question understanding through adaptive reasoning instructions using \model{GPT-3.5-Turbo-0613};
TempAgent~\cite{qianyihu-etal-2025-TempAgent} adapts the \textit{ReAct} framework with temporal reasoning toolkits powered by \model{GPT-4-Turbo};
MemoTime~\cite{tan2025memotime} retrieves past solution chains as external memory to enhance reasoning consistency, supporting multiple APIs including \model{GPT-4o-mini}, \model{GPT-4o}, \model{DeepSeek-V3}, and \model{DeepSeek-R1}. For fair comparison, we report results using \model{DeepSeek-V3} in our main experiments;
RTQA~\cite{gong-etal-2025-rtqa} decomposes complex temporal questions into sequential sub-tasks solved recursively using \model{GPT-4o-mini} combined with \model{DeepSeek-V3}.

\noindent\textbf{(3) LLM with Fine-Tuning.}
Instead of prompt engineering, these approaches fine-tune open-source language models to enhance temporal reasoning capabilities, though they still incorporate API calls or follow rigid predefined pipelines.
Search-R1~\cite{jin2025search-R1} applies zero-shot reinforcement learning to teach the \model{Qwen2.5-7B} model the capability of deciding how and when to call external search tools;
TimeR$^{4}$~\cite{qian2024timer4} explicitly exposes hidden temporal constraints through question rewriting and self-consistency filtering. Its pipeline includes a fine-tuned SentenceBERT retriever, \model{GPT-3.5-Turbo-0125} for rewriting, and a fine-tuned \model{LLaMA2-Chat-7B} generator;
PoK~\cite{qian2025poK} extends the rewriting strategy by introducing planning tokens that guide step-wise execution. It employs a fine-tuned \model{Qwen3-Embedding-0.7B} retriever, \model{GPT-4o} for planning, and a fine-tuned \model{LLaMA2-Chat-7B} generator.

\noindent\textbf{Evaluation on Complex \textsc{TimelineKGQA} Datasets.}
For the more challenging \textsc{Timeline-ICEWS-Actor} and \textsc{Timeline-CronQuestion} datasets, which require sophisticated temporal interval reasoning and duration computation, existing embedding-based models are not directly applicable. Therefore, we adopt five strong neural and LLM-based baselines representing the current state of the art in timeline reasoning: RTQA and PoK. For the remaining three baselines, we report results directly from the PoK paper~\cite{qian2025poK}.

\section{Setup Details}
\label{app:details}
This section outlines the configurations for the SFT and GRPO training stages.

\paragraph{Supervised Fine‑Tuning (SFT).}
We fine-tune \model{LLaMA3.1-8B-Instruct} using LLaMA-Factory for 2 epochs with a learning rate of $2\times10^{-5}$ and a batch size of 16. A key technical detail is the masking of $\langle \text{information} \rangle$ blocks to zero-out their loss, ensuring the model focuses on policy learning rather than redundant retrieval generation. Training employs \texttt{bf16} precision and AdamW optimizer with a 0.1 warmup ratio.

\paragraph{Group Relative Policy Optimization (GRPO).}
Post-SFT, reinforcement learning is conducted via a reverse curriculum on 3$\times$A800 GPUs. We employ vLLM for rollouts with a temperature of 1.0 and a group size of 5. For GRPO, we set the clipping threshold $\epsilon=0.2$ and KL penalty $\beta=0.01$. The learning rate for the actor is $5\times10^{-7}$. FSDP and gradient clipping (5.0) are applied to ensure training stability.

\begin{table*}[t]

\centering
\small
\setlength{\tabcolsep}{5pt}
\begin{tabular}{lcc}
\toprule
 & \textbf{Temp-R1 (Ours)} & \textbf{TimeR$^{4}$} \\
\midrule
Training Paradigm 
& RL (GRPO) + minimal SFT 
& Fully Supervised \\

Model Components 
& 1 (Unified agent) 
& $\geq$3 (Retriever + LLM + Rewrite) \\
\midrule
SFT Data Size 
& 1,000 
& $\sim$75,357 (20\% MULTITQ) \\

SFT Epochs 
& 2 
& 2 \\
\midrule
Main Optimization Data 
& $\sim$35,100 rollout trajectories 
& $\sim$150,714 supervised instances \\
\midrule
Retriever Training 
& --
& Contrastive learning \\

Retriever Epochs 
& -- 
& 10 \\

Negative Sampling 
& -- 
& 1 in-batch + 3 hard negatives \\
\midrule
External LLM Cost 
& $\sim$\$50 (one-time, SFT construction) 
& Continuous (GPT-3.5-turbo API call) \\

\bottomrule
\end{tabular}
\caption{\textbf{Comparison of Training Cost and Supervision Scale} between Temp-R1 and TimeR$^{4}$.}
\label{tab:training_cost}
\end{table*}

\begin{table*}[t]
\centering
\small
\setlength{\tabcolsep}{2.5pt}
\begin{tabular}{lcccc}
\toprule
\textbf{Model} & \textbf{Paradigm} & \textbf{LLM Access} & \textbf{API Calls} & \textbf{Inference Cost} \\
\midrule
TempAgent & Prompt-Engineering & Closed-source (GPT-4-turbo) & Multiple / query & High \\
RTQA & Prompt-Engineering & Closed-source (GPT-4o-mini + DeepSeek-V3) & Multiple / query & High \\
MemoTime & Prompt-Engineering & Closed-source (DeepSeek-V3) & Multiple / query & High \\
\midrule
TimeR$^{4}$ & Fine-tuning + Prompting & Hybrid (GPT-3.5-turbo + LLaMA2) & Per query rewrite & Medium \\
PoK & Fine-tuning + Prompting & Hybrid (GPT-4o + LLaMA2) & Per query plan & Medium \\
\midrule
\textbf{Temp-R1} & RL-trained Agent & Open-source (LLaMA3.1-8B) & None & Low \\
\bottomrule
\end{tabular}
\caption{\textbf{Inference-Time Cost Comparison} on MultiTQ. Prompt-based methods rely on closed-source LLM APIs during inference, while Temp-R1 performs fully local inference after training.}
\label{tab:inference_cost}
\end{table*}

\section{Efficiency Analysis}
\label{app:efficiency}
\subsection{Training Cost Analysis}

Table~\ref{tab:training_cost} compares the training cost of Temp-R1 and TimeR$^{4}$. Temp-R1 employs a unified agent trained primarily via reinforcement learning, requiring only 1,000 supervised instances for warm-up and approximately 35k rollout trajectories for optimization. In contrast, TimeR$^{4}$ relies on multi-stage supervised training with over 150k labeled instances and a separately trained retriever. Moreover, TimeR$^{4}$ depends on continuous external \texttt{GPT-3.5-turbo} API calls for question rewriting, introducing additional inference-time costs.

Overall, Temp-R1 achieves competitive performance with substantially fewer supervised signals and a simpler training pipeline, demonstrating improved training efficiency.

\subsection{Inference-Time Cost Analysis}

We further analyze inference-time efficiency by comparing Temp-R1 with representative prompt-based methods that rely on closed-source large language models. As summarized in Table~\ref{tab:inference_cost}, prompt-engineering approaches such as TempAgent~\cite{qianyihu-etal-2025-TempAgent}, RTQA~\cite{gong-etal-2025-rtqa}, and MemoTime~\cite{tan2025memotime} achieve strong performance but require multiple API calls to proprietary models during inference, resulting in continuous and non-negligible deployment costs.

In contrast, Temp-R1 is fully trained offline and performs inference using a single open-source model without any external API dependency. Once training is completed, no additional cost is incurred beyond standard model inference. This makes Temp-R1 more suitable for large-scale or long-term deployment scenarios where inference efficiency and cost stability are critical.

\definecolor{thinkblue}{RGB}{0, 0, 255}
\definecolor{searchcyan}{RGB}{0, 176, 240}
\definecolor{infobrown}{RGB}{192, 112, 0}
\definecolor{answerred}{RGB}{204, 0, 0}
\definecolor{planpurple}{RGB}{147, 39, 143}  
\definecolor{filterorange}{RGB}{247, 147, 30} 
\definecolor{rankgreen}{RGB}{0, 146, 69}     

\section{Case Study}
\label{app:case study}

This section provides a detailed examination of the reasoning trajectories generated by \textbf{Temp-R1} across the six temporal question types in the \textsc{MultiTQ} dataset. 
For comparative temporal questions such as the ``Before/After'' type (Table~\ref{tab:case_study_before/after}), the trajectories demonstrate the model's ability to apply strict temporal filters to isolate events within a specified timeframe. 
In ``Before-last'' sequential reasoning scenarios (Table~\ref{tab:case_study_before-last}), the model successfully executes a multi-step logic by first identifying a reference event's date and then performing a secondary search to locate the most recent preceding occurrence. 
Similarly, for ``First/Last'' queries (Table~\ref{tab:case_study_first/last}), the model utilizes its ranking and chronological sorting modules to compare multiple retrieved candidates and identify the chronological extreme.

The model also demonstrates high precision in handling varied temporal constraints and anchoring events. For simple ``Equal'' factual questions (Table~\ref{tab:case_study_equal}), the trajectory shows direct mapping between specific entities and their corresponding dates. 
In the more challenging ``Multi-equal'' multi-hop scenarios (Table~\ref{tab:case_study_multi-equal}), \texttt{Temp-R1} performs effective temporal bridging, using the time window of a pivot event to anchor the search for concurrent activities involving different entities. 
Finally, for ``After-first'' relative sequences (Table~\ref{tab:case_study_after-first}), the reasoning path highlights the model's ability to establish a historical anchor and chronologically evaluate subsequent events to find the immediate successor. 
Together, these trajectories confirm the robustness and structural clarity of \textbf{Temp-R1} in resolving multiple temporal granularities and multiple constraints.


\begin{table*}[ht]
\centering
\small
\begin{tabularx}{\textwidth}{X}
\toprule
\textbf{Question:} After May 2015, with which country did Oman formally sign an agreement? \\ \midrule
\textbf{Ground Truth:} ['Japan' 'Malaysia' 'Iran'] \\ \midrule
\textbf{\textit{Temp-R1}}: \\
{\color{planpurple}\texttt{<plan>}} \\
- Question type: Comparative temporal \\
- Time constraints: "After May 2015" (temporal scope) \\
- Sub-questions: \\
\quad 1. Identify agreements signed by Oman after May 2015. \\
\quad 2. Determine which countries were involved. \\
- Answer format: Country name \\
{\color{planpurple}\texttt{</plan>}} \\

{\color{searchcyan}\texttt{<search>}} Oman agreement signed after May 2015 {\color{searchcyan}\texttt{</search>}} \\

{\color{infobrown}\texttt{<information>}} \\
Doc 1(Title: Oman Sign\_formal\_agreement Malaysia on 2015-10-22) \\
Doc 2(Title: Oman Sign\_formal\_agreement Kuwait on 2015-02-19) \\
Doc 3(Title: Oman Sign\_formal\_agreement Iran on 2015-05-26) \\
Doc 4(Title: Malaysia Sign\_formal\_agreement Oman on 2015-10-22) \\
Doc 5(Title: Kuwait Sign\_formal\_agreement Oman on 2015-02-19) \\
Doc 6(Title: Oman Sign\_formal\_agreement Japan on 2015-06-23) \\
Doc 7(Title: Iran Sign\_formal\_agreement Oman on 2015-05-26) \\
Doc 8(Title: Japan Sign\_formal\_agreement Oman on 2015-06-23) \\
Doc 9(Title: Oman Sign\_formal\_agreement Iran on 2015-09-21) \\
Doc 10(Title: Iran Sign\_formal\_agreement Oman on 2015-09-21) \\
Doc 11(Title: United\_Arab\_Emirates Sign\_formal\_agreement Oman on 2005-05-24) \\
Doc 12(Title: Oman Sign\_formal\_agreement United\_Arab\_Emirates on 2005-05-24) \\
... (omitted 18 Docs) \\
{\color{infobrown}\texttt{</information>}} \\

{\color{filterorange}\texttt{<filter>}} \\
Doc 6(Title: Oman Sign\_formal\_agreement Japan on 2015-06-23) \\
Doc 9(Title: Oman Sign\_formal\_agreement Iran on 2015-09-21) \\
{\color{filterorange}\texttt{</filter>}} \\

{\color{thinkblue}\texttt{<think>}} The filtered facts show two agreements signed by Oman after May 2015: with Japan on 2015-06-23 and Iran on 2015-09-21. {\color{thinkblue}\texttt{</think>}} \\

{\color{answerred}\texttt{<answer>}} Japan {\color{answerred}\texttt{</answer>}} \\
\bottomrule
\end{tabularx}
\caption{Case study using a before/after question.}
\label{tab:case_study_before/after}
\end{table*}

\begin{table*}[ht]
\centering
\small
\begin{tabularx}{\textwidth}{X}
\toprule
\textbf{Question: Who made the last request for Iraq before Médecins Sans Frontières?}  \\ \midrule
\textbf{Ground Truth: ['UN Security Council']}  \\ \midrule
\textbf{\textit{Temp-R1 Output:}} \\
{\color{planpurple}\texttt{<plan>}} \\
- Question type: Sequential temporal \\
- Time constraints: ``last'' (comparative - need most recent before another event) \\
- Sub-questions: \\
\quad 1. When did Médecins Sans Frontières make the request for Iraq? \\
\quad 2. Who made the request before that date? \\
\quad 3. Identify the most recent request before the one by Médecins Sans Frontières. \\
- Answer format: Entity name \\
{\color{planpurple}\texttt{</plan>}} \\

{\color{thinkblue}\texttt{<think>}} First, I need to find when the request for Iraq by Médecins Sans Frontières was made. {\color{thinkblue}\texttt{</think>}} \\

{\color{searchcyan}\texttt{<search>}} Médecins Sans Frontières request Iraq {\color{searchcyan}\texttt{</search>}} \\

{\color{infobrown}\texttt{<information>}} \\
Doc 1(Title: Doctors\_Without\_Borders Make\_an\_appeal\_or\_request Iraq on 2014-06-18) \\
Doc 2(Title: Doctors\_Without\_Borders Make\_an\_appeal\_or\_request Niger on 2008-10-21) \\
Doc 3(Title: Reporters\_Sans\_Frontieres Make\_an\_appeal\_or\_request Interim\_Government\_(Iraq) on 2005-01-28) \\
Doc 4(Title: France Make\_an\_appeal\_or\_request Iraq on 2007-01-14) \\
Doc 5(Title: Doctors\_Without\_Borders Make\_an\_appeal\_or\_request Other\_Authorities\_/\_Officials\_(Chad) on 2014-04-03) \\
Doc 6(Title: International\_Government\_Organizations Make\_an\_appeal\_or\_request Iraq on 2007-02-18) \\
Doc 7(Title: France Make\_an\_appeal\_or\_request Iraq on 2006-11-06) \\
Doc 8(Title: France Make\_an\_appeal\_or\_request Iraq on 2006-01-29) \\
Doc 9(Title: Doctors\_Without\_Borders Make\_an\_appeal\_or\_request Other\_Authorities\_/\_Officials\_(Chad) on 2014-04-02) \\
Doc 10(Title: France Make\_an\_appeal\_or\_request Iraq on 2014-08-13) \\
... (omitted 20 Docs) \\
{\color{infobrown}\texttt{</information>}} \\

{\color{thinkblue}\texttt{<think>}} The request for Iraq by the ``Doctors\_Without\_Borders'' was on 2014-06-18. Now, I need to find who made the most recent request before this date. {\color{thinkblue}\texttt{</think>}} \\

{\color{searchcyan}\texttt{<search>}} request to Iraq before 2014-06-18 {\color{searchcyan}\texttt{</search>}} \\

{\color{infobrown}\texttt{<information>}} \\
Doc 1(Title: UN\_Security\_Council Make\_an\_appeal\_or\_request Iraq on 2014-06-06) \\
Doc 2(Title: Barack\_Obama Make\_an\_appeal\_or\_request Iraq on 2014-06-24) \\
Doc 3(Title: Barack\_Obama Make\_an\_appeal\_or\_request Iraq on 2014-06-23) \\
Doc 4(Title: Barack\_Obama Make\_an\_appeal\_or\_request Iraq on 2014-06-19) \\
Doc 5(Title: Barack\_Obama Make\_an\_appeal\_or\_request Iraq on 2014-06-21) \\
Doc 6(Title: Nuri\_al-Maliki Make\_an\_appeal\_or\_request Iraq on 2014-06-25) \\
Doc 7(Title: UN\_Security\_Council Make\_an\_appeal\_or\_request Iraq on 2014-05-01) \\
Doc 8(Title: Iraq Make\_an\_appeal\_or\_request Barack\_Obama on 2014-06-12) \\
Doc 9(Title: Barack\_Obama Make\_an\_appeal\_or\_request Iraq on 2014-07-02) \\
Doc 10(Title: Barack\_Obama Demand Iraq on 2014-06-19) \\
... (omitted 20 Docs) \\
{\color{infobrown}\texttt{</information>}} \\

{\color{filterorange}\texttt{<filter>}} \\
Doc 1 (Title: UN\_Security\_Council Make\_an\_appeal\_or\_request Iraq on 2014-06-06) \\
{\color{filterorange}\texttt{</filter>}} \\

{\color{thinkblue}\texttt{<think>}} The most recent request before the one by the ``Doctors\_Without\_Borders'' on 2014-06-18 was by the UN Security Council on 2014-06-06. {\color{thinkblue}\texttt{</think>}} \\

{\color{answerred}\texttt{<answer>}} UN Security Council {\color{answerred}\texttt{</answer>}} \\
\bottomrule
\end{tabularx}
\caption{Case study using a before-last question.}
\label{tab:case_study_before-last}
\end{table*}

\begin{table*}[ht]
\centering
\small
\begin{tabularx}{\textwidth}{X}
\toprule
\textbf{Question: In which year did the Chancellor of the Exchequer visit China for the first time?}  \\ \midrule
\textbf{Ground Truth: ['2005']}  \\ \midrule
\textbf{\textit{Temp-R1 Output:}} \\
{\color{planpurple}\texttt{<plan>}} \\
- Question type: Multi-hop \\
- Time constraints: ``first time'' (comparative - need earliest date) \\
- Sub-questions: \\
\quad 1. When did the Chancellor of the Exchequer visit China? \\
\quad 2. Identify the earliest such visit. \\
- Answer format: yyyy \\
{\color{planpurple}\texttt{</plan>}} \\

{\color{thinkblue}\texttt{<think>}} First, I need to find visits to China by the Chancellor of the Exchequer and identify the earliest one. {\color{thinkblue}\texttt{</think>}} \\

{\color{searchcyan}\texttt{<search>}} Chancellor of the Exchequer visit China {\color{searchcyan}\texttt{</search>}} \\

{\color{infobrown}\texttt{<information>}} \\
Doc 1(Title: Chancellor\_of\_the\_Exchequer Make\_a\_visit China on 2005-02-21) \\
Doc 2(Title: Chancellor\_of\_the\_Exchequer Make\_a\_visit China on 2015-09-22) \\
Doc 3(Title: China Host\_a\_visit Chancellor\_of\_the\_Exchequer on 2005-02-21) \\
Doc 4(Title: China Host\_a\_visit Chancellor\_of\_the\_Exchequer on 2015-09-22) \\
Doc 5(Title: Chancellor\_of\_the\_Exchequer Express\_intent\_to\_meet\_or\_negotiate China on 2012-01-14) \\
Doc 6(Title: Chancellor\_of\_the\_Exchequer Express\_intent\_to\_meet\_or\_negotiate China on 2012-01-13) \\
Doc 7(Title: Chancellor\_of\_the\_Exchequer Make\_statement China on 2005-02-21) \\
Doc 8(Title: Wen\_Jiabao Express\_intent\_to\_meet\_or\_negotiate Chancellor\_of\_the\_Exchequer on 2009-01-22) \\
Doc 9(Title: Wen\_Jiabao Consult Chancellor\_of\_the\_Exchequer on 2005-02-21) \\
Doc 10(Title: Chancellor\_of\_the\_Exchequer Consult Wen\_Jiabao on 2005-02-21) \\
... (omitted 20 Docs) \\
{\color{infobrown}\texttt{</information>}} \\

{\color{filterorange}\texttt{<filter>}} \\
Doc 1 (Chancellor of the Exchequer Make a visit China on 2005-02-21) \\
Doc 2 (Chancellor of the Exchequer Make a visit China on 2015-09-22) \\
{\color{filterorange}\texttt{</filter>}} \\

{\color{rankgreen}\texttt{<rank>}} \\
Doc 1 (Chancellor of the Exchequer Make a visit China on 2005-02-21) \\
Doc 2 (Chancellor of the Exchequer Make a visit China on 2015-09-22) \\
{\color{rankgreen}\texttt{</rank>}} \\

{\color{thinkblue}\texttt{<think>}} The earliest visit of the Chancellor of the Exchequer to China was on 2005-02-21. {\color{thinkblue}\texttt{</think>}} \\

{\color{answerred}\texttt{<answer>}} 2005 {\color{answerred}\texttt{</answer>}} \\
\bottomrule
\end{tabularx}
\caption{Case study using a first/last question.}
\label{tab:case_study_first/last}
\end{table*}

\begin{table*}[ht]
\centering
\small
\begin{tabularx}{\textwidth}{X}
\toprule
\textbf{Question: When did Mswati III denounce Swaziland?}  \\ \midrule
\textbf{Ground Truth: ['2011-03-19']}  \\ \midrule
\textbf{\textit{Temp-R1 Output:}} \\
{\color{planpurple}\texttt{<plan>}} \\
- Question type: Simple factual \\
- Time constraints: None explicitly stated \\
- Sub-questions: None needed \\
- Answer format: yyyy-mm-dd \\
{\color{planpurple}\texttt{</plan>}} \\

{\color{searchcyan}\texttt{<search>}} Mswati III denounce Swaziland {\color{searchcyan}\texttt{</search>}} \\

{\color{infobrown}\texttt{<information>}} \\
Doc 1(Title: Mswati\_III Criticize\_or\_denounce Swaziland on 2011-03-19) \\
Doc 2(Title: Swaziland Criticize\_or\_denounce Mswati\_III on 2011-04-12) \\
Doc 3(Title: Swaziland\_Solidarity\_Network Criticize\_or\_denounce Mswati\_III on 2011-03-05) \\
Doc 4(Title: Mswati\_III Reject Legislature\_(Swaziland) on 2012-11-15) \\
Doc 5(Title: Mswati\_III Make\_statement Swaziland on 2006-02-10) \\
Doc 6(Title: Mswati\_III Make\_statement Swaziland on 2007-09-29) \\
Doc 7(Title: Mswati\_III Make\_statement Swaziland on 2005-08-16) \\
Doc 8(Title: Mswati\_III Make\_statement Swaziland on 2008-09-05) \\
Doc 9(Title: Mswati\_III Make\_statement Swaziland on 2011-09-14) \\
Doc 10(Title: Mswati\_III Make\_statement Swaziland on 2012-07-25) \\
... (omitted 20 Docs) \\
{\color{infobrown}\texttt{</information>}} \\

{\color{filterorange}\texttt{<filter>}} \\
Doc 1: Mswati\_III Criticize\_or\_denounce Swaziland on 2011-03-19 \\
{\color{filterorange}\texttt{</filter>}} \\

{\color{answerred}\texttt{<answer>}} 2011-03-19 {\color{answerred}\texttt{</answer>}} \\
\bottomrule
\end{tabularx}
\caption{Case study using a equal question.}
\label{tab:case_study_equal}
\end{table*}

\begin{table*}[ht]
\centering
\small
\begin{tabularx}{\textwidth}{X}
\toprule
\textbf{Question: Who was Kuwait negotiating with in the same month as Christine Lagarde?}  \\ \midrule
\textbf{Ground Truth: ['UN Security Council']}  \\ \midrule
\textbf{\textit{Temp-R1 Output:}} \\
{\color{planpurple}\texttt{<plan>}} \\
- Question type: Multi-hop \\
- Time constraints: ``in the same month as Christine Lagarde'' (temporal scope) \\
- Sub-questions: \\
\quad 1. When did Christine Lagarde have relevant negotiations or event with Kuwait? \\
\quad 2. Who was Kuwait negotiating with in that same month? \\
- Answer format: Entity name or concise phrase \\
{\color{planpurple}\texttt{</plan>}} \\

{\color{thinkblue}\texttt{<think>}} First, I need to find out when Christine Lagarde had relevant negotiations or event with Kuwait. {\color{thinkblue}\texttt{</think>}} \\

{\color{searchcyan}\texttt{<search>}} Christine Lagarde negotiations Kuwait {\color{searchcyan}\texttt{</search>}} \\

{\color{infobrown}\texttt{<information>}} \\
Doc 1(Title: Christine\_Lagarde Engage\_in\_negotiation Kuwait on 2009-06-21) \\
Doc 2(Title: Kuwait Engage\_in\_negotiation Christine\_Lagarde on 2009-06-21) \\
Doc 3(Title: Christine\_Lagarde Consider\_policy\_option Kuwait on 2015-11-11) \\
Doc 4(Title: Christine\_Lagarde Engage\_in\_negotiation Abdel\_Fattah\_Al-Sisi on 2014-06-07) \\
Doc 5(Title: Christine\_Lagarde Make\_statement Kuwait on 2009-06-22) \\
Doc 6(Title: Abdel\_Fattah\_Al-Sisi Engage\_in\_negotiation Christine\_Lagarde on 2014-06-07) \\
Doc 7(Title: Christine\_Lagarde Express\_intent\_to\_engage\_in\_diplomatic\_cooperation\_(such\_as\_policy\_support) UAE on 2006-11-27) \\
Doc 8(Title: Christine\_Lagarde Express\_intent\_to\_meet\_or\_negotiate Middle\_East on 2014-05-06) \\
Doc 9(Title: Christine\_Lagarde Engage\_in\_negotiation Yannis\_Stournaras on 2014-06-05) \\
Doc 10(Title: Christine\_Lagarde Express\_intent\_to\_meet\_or\_negotiate Central\_Bank\_(Jordan) on 2014-05-06) \\
... (omitted 20 Docs) \\
{\color{infobrown}\texttt{</information>}} \\

{\color{thinkblue}\texttt{<think>}} The relevant negotiation or event with Christine Lagarde and Kuwait occurred on 2009-06-21. Now I need to find out who Kuwait was negotiating with in June 2009. {\color{thinkblue}\texttt{</think>}} \\

{\color{searchcyan}\texttt{<search>}} Kuwait negotiation June 2009 {\color{searchcyan}\texttt{</search>}} \\

{\color{infobrown}\texttt{<information>}} \\
Doc 1(Title: Kuwait Engage\_in\_negotiation Iraq on 2009-07-27) \\
Doc 2(Title: Kuwait Engage\_in\_negotiation Oman on 2008-06-10) \\
Doc 3(Title: UN\_Security\_Council Engage\_in\_negotiation Kuwait on 2009-06-03) \\
Doc 4(Title: UAE Engage\_in\_negotiation Kuwait on 2008-06-10) \\
Doc 5(Title: Oman Engage\_in\_negotiation Kuwait on 2008-06-10) \\
Doc 6(Title: Iraq Engage\_in\_negotiation Kuwait on 2009-07-27) \\
Doc 7(Title: Kuwait Engage\_in\_negotiation UN\_Security\_Council on 2009-06-03) \\
Doc 8(Title: Kuwait Engage\_in\_negotiation UAE on 2008-06-10) \\
Doc 9(Title: Kuwait Engage\_in\_negotiation Christine\_Lagarde on 2009-06-21) \\
Doc 10(Title: Christine\_Lagarde Engage\_in\_negotiation Kuwait on 2009-06-21) \\
... (omitted 20 Docs) \\
{\color{infobrown}\texttt{</information>}} \\

{\color{filterorange}\texttt{<filter>}} \\
Doc 9(Title: Kuwait Engage\_in\_negotiation Christine\_Lagarde on 2009-06-21) \\
Doc 7(Title: Kuwait Engage\_in\_negotiation UN\_Security\_Council on 2009-06-03) \\
{\color{filterorange}\texttt{</filter>}} \\

{\color{thinkblue}\texttt{<think>}} The filtered facts show that in June 2009, besides Christine Lagarde, the UN Security Council was involved in negotiations with Kuwait. {\color{thinkblue}\texttt{</think>}} \\

{\color{answerred}\texttt{<answer>}} UN Security Council {\color{answerred}\texttt{</answer>}} \\
\bottomrule
\end{tabularx}
\caption{Case study using a multi-equal question.}
\label{tab:case_study_multi-equal}
\end{table*}

\begin{table*}[ht]
\centering
\small
\begin{tabularx}{\textwidth}{X}
\toprule
\textbf{Question: After Ethiopia, against which country did South Sudan first use conventional military force?}  \\ \midrule
\textbf{Ground Truth: ['Sudan']}  \\ \midrule
\textbf{\textit{Temp-R1 Output:}} \\
{\color{planpurple}\texttt{<plan>}} \\
- Question type: Sequential temporal \\
- Time constraints: ``first use after Ethiopia'' (sequential - need earliest after a specific event) \\
- Sub-questions: \\
\quad 1. When did South Sudan use conventional military force against Ethiopia? \\
\quad 2. What other countries has South Sudan used conventional military force against, and when? \\
\quad 3. Among these events, which occurred first after the one against Ethiopia? \\
- Answer format: Country name \\
{\color{planpurple}\texttt{</plan>}} \\

{\color{thinkblue}\texttt{<think>}} First, I need to find when South Sudan used conventional military force against Ethiopia. {\color{thinkblue}\texttt{</think>}} \\

{\color{searchcyan}\texttt{<search>}} South Sudan use conventional military force Ethiopia {\color{searchcyan}\texttt{</search>}} \\

{\color{infobrown}\texttt{<information>}} \\
Doc 1(Title: South\_Sudan Use\_conventional\_military\_force Ethiopia on 2012-01-06) \\
Doc 2(Title: South\_Sudan Use\_conventional\_military\_force Combatant\_(Lord's\_Resistance\_Army) on 2008-06-16) \\
Doc 3(Title: South\_Sudan Use\_conventional\_military\_force Combatant\_(Lord's\_Resistance\_Army) on 2008-06-17) \\
Doc 4(Title: Sudan\_People's\_Liberation\_Army Use\_conventional\_military\_force South\_Sudan on 2011-02-14) \\
Doc 5(Title: Sudan\_People's\_Liberation\_Army Use\_conventional\_military\_force South\_Sudan on 2014-01-30) \\
Doc 6(Title: Sudan\_People's\_Liberation\_Army Use\_conventional\_military\_force Military\_(South\_Sudan) on 2012-04-14) \\
Doc 7(Title: Sudan\_People's\_Liberation\_Army Use\_conventional\_military\_force South\_Sudan on 2009-06-14) \\
Doc 8(Title: Sudan\_People's\_Liberation\_Army Use\_conventional\_military\_force South\_Sudan on 2014-02-02) \\
Doc 9(Title: Sudan\_People's\_Liberation\_Army Use\_conventional\_military\_force Military\_(South\_Sudan) on 2013-01-14) \\
Doc 10(Title: Sudan\_People's\_Liberation\_Army Use\_conventional\_military\_force Military\_(South\_Sudan) on 2013-01-29) \\
... (omitted 20 Docs) \\
{\color{infobrown}\texttt{</information>}} \\

{\color{thinkblue}\texttt{<think>}} The date when South Sudan used conventional military force against Ethiopia was 2012-01-06. Now, I need to find which country South Sudan used conventional military force against after this date and identify the earliest occurrence. {\color{thinkblue}\texttt{</think>}} \\

{\color{searchcyan}\texttt{<search>}} South Sudan use conventional military force after 2012-01-06 {\color{searchcyan}\texttt{</search>}} \\

{\color{infobrown}\texttt{<information>}} \\
Doc 1(Title: Military\_(South\_Sudan) Use\_conventional\_military\_force South\_Sudan on 2012-01-04) \\
Doc 2(Title: Sudan\_People's\_Liberation\_Army Use\_conventional\_military\_force Military\_(South\_Sudan) on 2012-04-14) \\
Doc 3(Title: Sudan\_People's\_Liberation\_Army Use\_conventional\_military\_force Military\_(South\_Sudan) on 2012-04-23) \\
Doc 4(Title: Military\_(South\_Sudan) Use\_conventional\_military\_force South\_Sudan on 2012-08-24) \\
Doc 5(Title: Military\_(South\_Sudan) Use\_conventional\_military\_force South\_Sudan on 2012-08-25) \\
Doc 6(Title: South\_Sudan Use\_conventional\_military\_force Sudan on 2012-02-26) \\
Doc 7(Title: Government\_(Sudan) Use\_conventional\_military\_force South\_Sudan on 2012-05-23) \\
Doc 8(Title: Military\_(South\_Sudan) Use\_conventional\_military\_force Indigenous\_People\_(South\_Sudan) on 2012-05-24) \\
Doc 9(Title: South\_Sudan Use\_conventional\_military\_force Ethiopia on 2012-01-06) \\
Doc 10(Title: South\_Sudan Use\_conventional\_military\_force Sudan on 2012-02-29) \\
... (omitted 20 Docs) \\
{\color{infobrown}\texttt{</information>}} \\

{\color{filterorange}\texttt{<filter>}} \\
Doc 10(Title: South\_Sudan Use\_conventional\_military\_force Sudan on 2012-02-29) \\
Doc 6(Title: South\_Sudan Use\_conventional\_military\_force Sudan on 2012-02-26) \\
{\color{filterorange}\texttt{</filter>}} \\

{\color{rankgreen}\texttt{<rank>}} \\
Doc 6(Title: South\_Sudan Use\_conventional\_military\_force Sudan on 2012-02-26) \\
Doc 10(Title: South\_Sudan Use\_conventional\_military\_force Sudan on 2012-02-29) \\
{\color{rankgreen}\texttt{</rank>}} \\

{\color{thinkblue}\texttt{<think>}} The earliest use of South Sudan's conventional military force after the one against Ethiopia on 2012-01-06 is against Sudan on 2012-02-26. {\color{thinkblue}\texttt{</think>}} \\

{\color{answerred}\texttt{<answer>}} Sudan {\color{answerred}\texttt{</answer>}} \\
\bottomrule
\end{tabularx}
\caption{Case study using a after-first question.}
\label{tab:case_study_after-first}
\end{table*}

\end{document}